\documentclass[runningheads]{llncs}
    \makeatletter
\def\@fnsymbol#1{\ensuremath{\ifcase#1\or \dagger\or \ddagger\or
   \mathsection\or \mathparagraph\or \|\or **\or \dagger\dagger
   \or \ddagger\ddagger \else\@ctrerr\fi}}
    \makeatother
 
\usepackage{eccv}

\usepackage{eccvabbrv}
\usepackage{bm}

\usepackage{graphicx}
\usepackage{booktabs}
\usepackage{afterpage}
\usepackage{multicol}
\usepackage{adjustbox} 
\usepackage{xcolor} 
\usepackage{wrapfig} 

\usepackage{soul}
\usepackage{pifont}
\usepackage{colortbl}
\usepackage{mathtools} 

\newcommand{\cmark}{\ding{51}}%

\usepackage[accsupp]{axessibility}  

\usepackage{amsmath} 
\usepackage{algorithm} 
\usepackage{algpseudocode} 

\usepackage{hyperref}

\usepackage{orcidlink}
\usepackage{algorithmicx}
\usepackage[capitalize]{cleveref}
\crefname{section}{Sec.}{Secs.}
\Crefname{section}{Section}{Sections}
\crefname{table}{Tab.}{Tabs.}
\Crefname{table}{Table}{Tables}

\begin{document}

\title{Diffusion Model for Robust Multi-Sensor Fusion in 3D Object Detection and BEV Segmentation}

\titlerunning{DifFUSER}


\author{
Duy-Tho Le\textsuperscript{1}\orcidlink{0000-0003-2356-4530}
\and Hengcan Shi\textsuperscript{2}\thanks{Corresponding author}\orcidlink{0000-0002-1340-0009}
\and Jianfei Cai\textsuperscript{1}\orcidlink{0000-0002-9444-3763}
\and Hamid Rezatofighi\textsuperscript{1}\orcidlink{0000-0002-8659-8773} \vspace{-0.7em}
}

\authorrunning{DT. Le et al.}

\institute{
\textsuperscript{1}Monash University, \textsuperscript{2} Hunan University \\ 
\email{tho.le1@monash.edu hengcan.shi@gmail.com}\\
\url{https://ldtho.github.io/DifFUSER/} \\
\vspace{-1.5em}
}

\maketitle


\begin{figure}[!htb]
    \vspace{-2em}
    \centering
    \includegraphics[width=\linewidth]{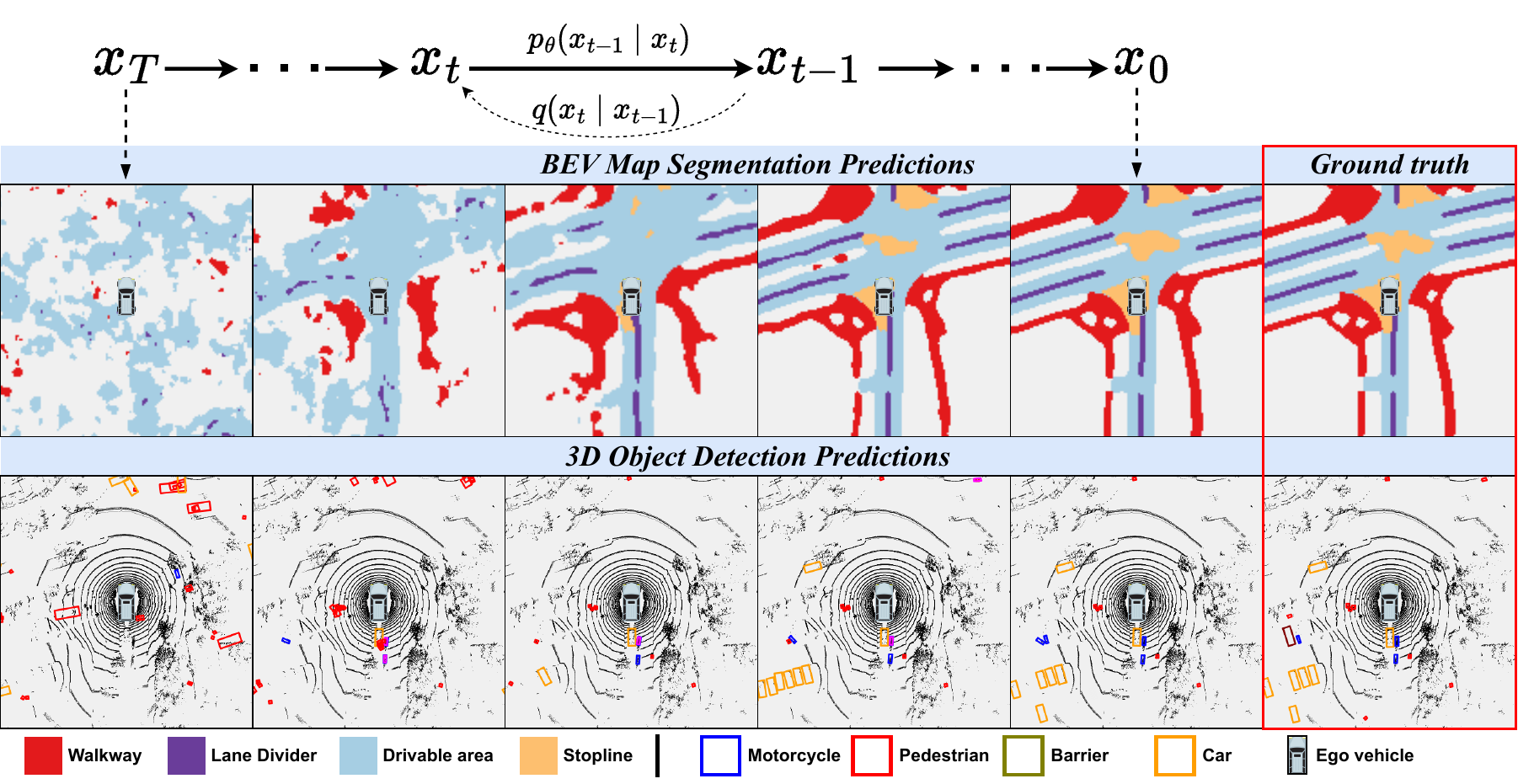}
    \vspace{-1em}
    \caption{We propose to use denoising diffusion process for multi-modal BEV features fusion, which are then used for 3D object detection and BEV map segmentation.}
    \label{fig:intro_comparison}
    \vspace{-3em}
\end{figure}

\begin{abstract}
  Diffusion models have recently gained prominence as powerful deep generative models, demonstrating unmatched performance across various domains. However, their potential in multi-sensor fusion remains largely unexplored. In this work, we introduce ``DifFUSER'', a novel approach that leverages diffusion models for multi-modal fusion in 3D object detection and BEV map segmentation. Benefiting from the inherent denoising property of diffusion, DifFUSER is able to refine or even synthesize sensor features in case of sensor malfunction, thereby improving the quality of the fused output. In terms of architecture, our DifFUSER blocks are chained together in a hierarchical BiFPN fashion, termed cMini-BiFPN, offering an alternative architecture for latent diffusion. We further introduce a Gated Self-conditioned Modulated (GSM) latent diffusion module together with a Progressive Sensor Dropout Training (PSDT) paradigm, designed to add stronger conditioning to the diffusion process and robustness to sensor failures. Our extensive evaluations on the Nuscenes dataset reveal that DifFUSER not only achieves state-of-the-art performance with a 70.04\% mIOU in BEV map segmentation tasks but also competes effectively with leading transformer-based fusion techniques in 3D object detection. 
  \keywords{3D Object Detection \and BEV Map Segmentation \and Diffusion}
\end{abstract}

\section{Introduction}
3D object detection and Bird's Eye View (BEV) map segmentation are fundamental tasks in autonomous driving. The former involves detecting objects in 3D space, while the latter segments BEV maps into semantic categories. A growing trend in this field is the fusion of features from multiple sensors, leveraging their complementary strengths. For example, 3D point clouds provide geometric data for 3D object detection, but lack the rich color information crucial for BEV semantic segmentation —a gap effectively filled by 2D images.

Recent transformer-based fusion methods \cite{deepinteraction, transfusion, cmt} have set new benchmarks in 3D object detection by learning feature mappings from both sensors through cross-attention mechanisms. Nevertheless, these methods are less adaptable for extra tasks like BEV map segmentation. An alternative is to create unified BEV representations from both sensors, as seen in LSS-based \cite{lss} approaches. However, these often yield sub-optimal results due to: 1) inadequately designed fusion module architectures that fail to capture the intricate relationship between the sensors, and 2) the inherent noise between different modalities, causing the fused features to be noisy and inaccurate.

To address these challenges, we explore the use of generative models, particularly Diffusion Probabilistic Models (DPMs), for multi-modal fusion and denoising. DPMs have shown promise in various applications, including image generation \cite{ddpm, ddim} and object detection \cite{diffusiondet}. Yet, the potential of DPMs in multi-sensor fusion, especially between 3D point clouds and 2D multi-view images, remains unexplored. To this end, we propose \emph{DifFUSER}, a conditional diffusion-based generative model with enhanced fusion architecture designed for multi-modal learning in 3D perception. It processes features from both sensors to output a refined BEV representation, which is subsequently directed to task-specific heads for 3D object detection and BEV map segmentation. We also show DifFUSER's robustness in synthesizing new features to compensate for missing modality, minimising performance loss and ensuring reliability even with compromised sensor.

In terms of architecture, our DifFUSER blocks are chained together in a hierarchical BiFPN (Bidirectional Feature Pyramid Network) fashion \cite{efficientdet, pifenet}, termed cMini-BiFPN, offering an alternative architecture for latent diffusion, particularly good in handling detailed and multi-scales features from different sensors. This architectural change is complemented by a Gated Self-conditioned Modulated (GSM) latent diffusion module, designed to strengthen the conditioning modulation during diffusion, leading to more precise denoising and feature enhancement. We train DifFUSER end-to-end using our proposed Progressive Sensor Dropout Training (PSDT) paradigm, together with task-specific losses, which enhance the model's robustness and capability in denoising corrupted BEV features and generating more accurate outputs for downstream tasks.

\begin{wrapfigure}{r}{0.6\textwidth} 
  \centering
  \includegraphics[width=0.6\textwidth]{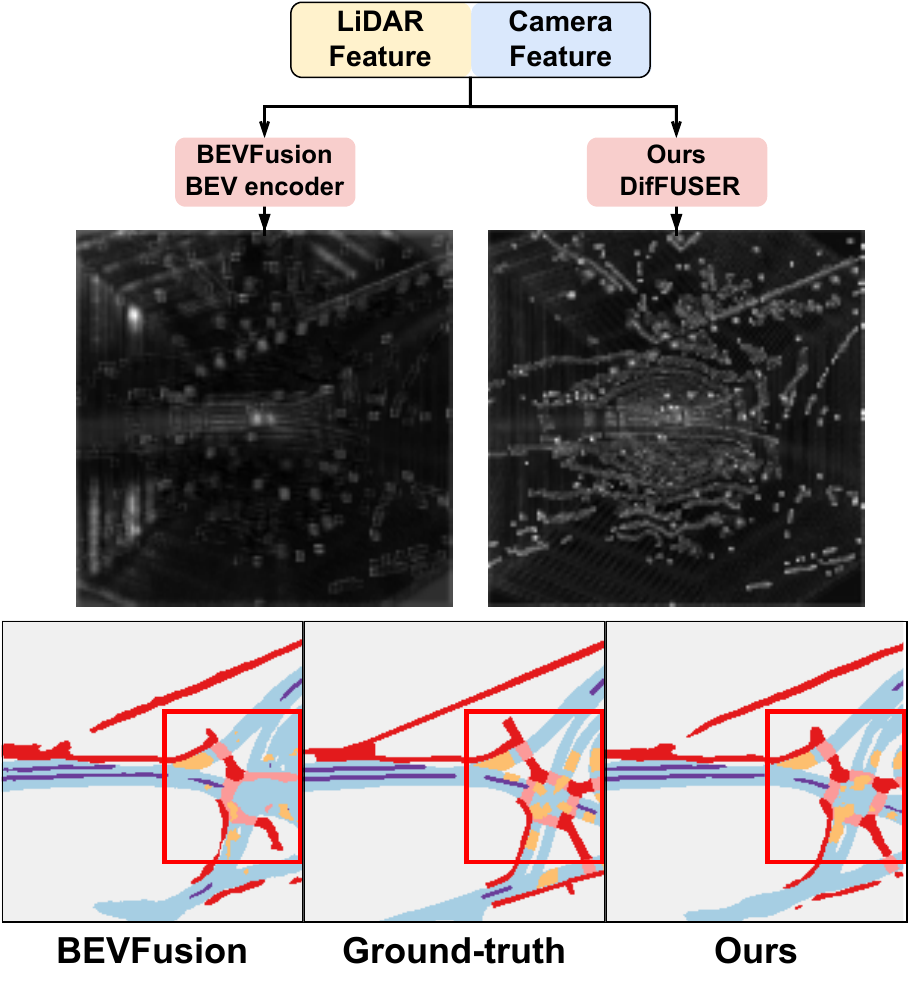} 
  \caption{Comparison of our DifFUSER fusion module with BEV encoder of the baseline. Our fusion module's output activation map is much more expressive than the baseline's, resulting in much better performance in downstream tasks.}
  \label{fig:intro_comparison}
  \vspace{-1.5em}
\end{wrapfigure}

\cref{fig:intro_comparison} presents the output activation maps from the BEV encoder of the baseline BEVFusion \cite{bevfusionmit} and our DifFUSER module. The activation map from our BEVFusion module is blurry, which might hinder the downstream task performance. In contrast, our fusion module, with better architecture design and the denoising property of the diffusion model, can effectively mitigate the noises and intrinsic differences between different sensors, notably improving the predictions and strong robustness against sensor malfunctioning. We conduct extensive experiments on the large-scale Nuscenes dataset \cite{nuscenes} to evaluate the performance and design choice of DifFUSER. The results show that DifFUSER achieves SOTA performance (70.04\% mIOU) in the BEV map segmentation task and performs on par with SOTA transformer-based fusion methods in the 3D object detection task. We also conduct ablation studies to show the effectiveness of each component in DifFUSER.

In summary, our contributions are listed as follows.
\begin{itemize}
    \item We are first to introduce a new DPM-based approach for multi-modal fusion in 3D perception. With a well-designed diffusion architecture (cMini-BiFPN) and training paradigm (PSDT), our method showcases enhanced multi-modal feature fusion and denoising capabilities, as well as robustness to sensor failure by generatively synthesizing missing sensor data.
    \item We propose a Gated Self-Conditioned Modulated (GSM) latent diffusion mechanism within the DifFUSER blocks, reinforcing the fusion process with stronger conditioning and shaping the diffusion trajectory towards high-quality features.  
    \item We conduct extensive experiments on the Nuscenes dataset \cite{nuscenes} to evaluate the performance of DifFUSER. The results show that DifFUSER surpasses the current SOTA by 7.34\% in BEV map segmentation tasks and performs on par with SOTA transformer-based fusion methods in 3D object detection.
\end{itemize}




\section{Related Work}

\textbf{Single-modality 3D perception.} Single-modality 3D perception has been widely studied in the past few years and mainly comprises two categories: 1) Lidar-based and 2) camera-based methods. Camera-based detection evolved from early dense prediction pipelines \cite{fcos3d, wang2022probabilistic} to sophisticated methods like DETR3D \cite{detr3d}, BEVFormer \cite{bevformer} and BEVDepth \cite{bevdepth}, which integrate Bird's Eye View (BEV) representations with transformer attention for multi-view fusion. In contrast, LiDAR-based detection relies on point cloud data, with pioneering architectures like PointNet \cite{pointnet} processing raw point clouds and subsequent developments like VoxelNet \cite{voxelnet} transforming into structured formats like voxels \cite{largekernel3d, second, focalconv}, range images \cite{rangedet, rsn}, and pillars \cite{pointpillars, tanet, pifenet}. These approaches reflect the ongoing evolution in object detection, leveraging the unique strengths of each modality.

\noindent\textbf{Multi-modality fusion approaches in 3D perception.} Fusion-based approaches have been widely adopted in 3D perception tasks and can be categorised into 3 main groups: 1) early fusion, 2) late fusion, and 3) feature fusion.
\textbf{\textit{Early fusion}} methods directly fuse the raw data from both sensors, either by augmenting the raw point cloud with image features \cite{pointpainting, fusionpainting, complexer-yolo} or leveraging the image features to guide the point cloud feature extraction \cite{f-pointnet, frustum-pointpillars}. The fusion is performed sequentially, introducing additional latency to the system and prone to error propagation.
\textbf{\textit{Late fusion}} methods \cite{fastclocs, clocs} operate on the predictions from both sensors, either by fusing the predictions \cite{fastclocs} or by fusing the features extracted from both sensors \cite{clocs}. The fusion is performed at the end of the pipeline, which is more efficient than early fusion methods but fails to capture the feature-to-feature relationships between both sensors. Thus the fusion results are often noisy and sub-optimal.
\textbf{\textit{Feature fusion}} methods \cite{bevfusionmit, deepinteraction, transfusion, cmt} have deeper feature fusion, where the features from both sensors are fused at multiple levels of the network. Although the fused representations are more expressive and yield better performance, LiDAR and camera features are intrinsically different and aligning their features is still a challenging task. Most current SOTA methods fall into this category. Particularly, \textit{transformer-based fusion} methods \cite{deepinteraction, transfusion, cmt} have been proposed to fuse the features from both sensors, leveraging feature-to-feature relationships captured by the transformer encoder-decoder architecture and achieving SOTA performance in 3D object detection. However, transformer-based fusion is usually computationally demanding and is not easy to adapt to other perception tasks such as BEV map segmentation. Another way of fusing multi-modalities is to construct an \textit{unified BEV representation} \cite{bevfusionmit, bevfusionpeking} from both sensors using LSS-based \cite{lss} backbone to lift the 2D image features to 3D space, and then apply multiple task-specific heads to the fused representation. This approach is more efficient and can be easily extended to multiple perception tasks. However, the fusion result is often sub-optimal due to the inherent noise and misalignment between different sensors. 

Our DifFUSER belongs to the feature-fusion sub-category, we aim to improve the fusion architecture and leverage the denoising property of generative models to mitigate the noises and intrinsical differences between different sensors, thus improving downstream tasks' performance.


\noindent\textbf{Generative models for 3D perception.} Generative models have been recently adopted in 3D perception tasks. However, most of them are for text-to-3D \cite{dreamfusion, magic3d}, image-to-3D \cite{zero123}, and 3D trajectory prediction \cite{motiondiffuser}. We notice that there are concurrent works \cite{diff3det, diffbev} that utilise diffusion models for 3D object detection, but they only operate on single-modality (either camera or LiDAR) and do not utilise diffusion for multi-modal fusion. 

In contrast, we aim to leverage the denoising property of generative models to mitigate the noises and intrinsic differences between different sensors, tackling some of the most fundamental tasks in autonomous driving, such as 3D object detection and BEV map segmentation. Our DifFUSER is the first to adapt diffusion models for multi-modal fusion for 3D perception tasks. The generated BEV feature can be shared and optimized end-to-end with downstream tasks.

\section{Methodology}
In this section, we introduce DifFUSER, a conditional generative model with enhanced fusion architecture, designed for multi-modal and multi-task learning within the scope of 3D object detection and BEV map segmentation. Our DifFUSER features take the extracted features from lidar and camera sensors and process them in three ways: one used as the diffusion target, one infused with Gaussian noise ($x^F_t$) and one is partially masked by PSDT ($\tilde{x}^F_0$). $\tilde{x}^F_0$ and $x^F_t$ are then input into the cMini-BiFPN (with GSM as encoder blocks) module, where they are denoised and refined. This enhanced representation is finally utilised by task-specific heads for precise 3D object detection and BEV map segmentation.


\begin{figure*}[!t]
    \centering
    \includegraphics[width=\linewidth]{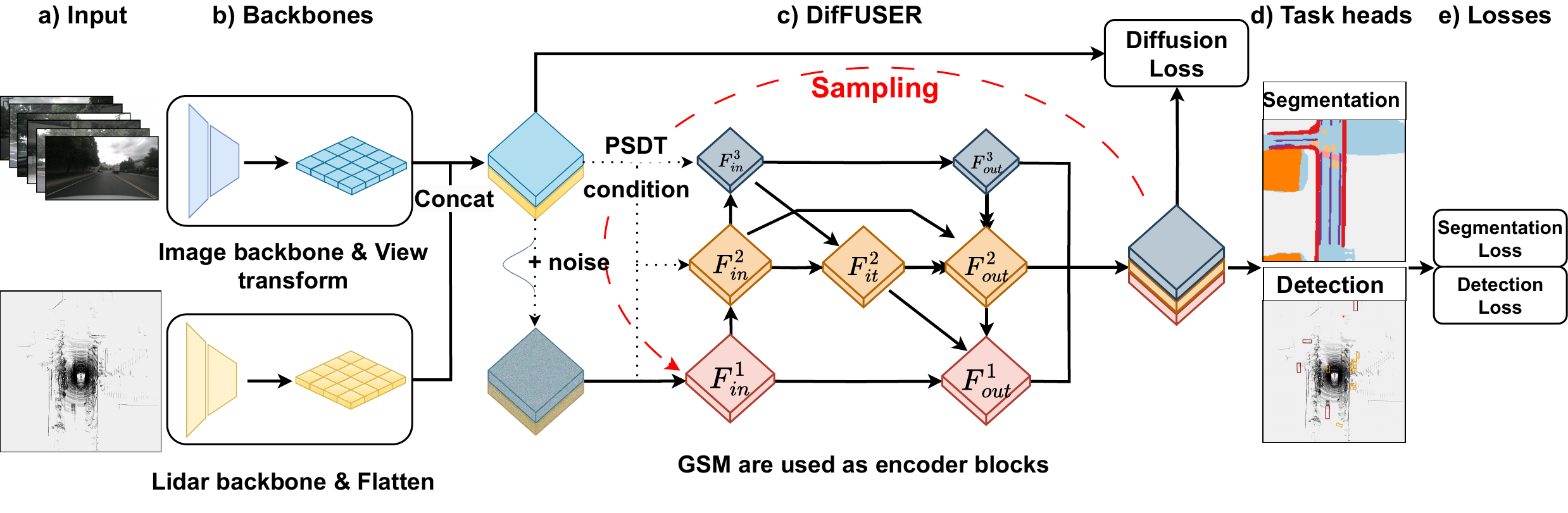}
    \caption{Our DifFUSER framework is structured to first process input data—comprising both point clouds and images—through respective backbones to create initial latent features. These features are then concatenated and fed into the DifFUSER blocks. Within these blocks, the concatenated feature is used as condition (partially masked out) to iteratively denoise the corrupted features, enhancing its quality at each step. The output feature is then used for downstream tasks.}
    \label{fig:Diffuser}
    \vspace{-1.5em}
\end{figure*}

\subsection{Fuser Architecture Design (cMini-BiFPN).} In autonomous driving, where fast processing is the top priority, the need to ensure both the performance and efficiency of the diffusion module is important. Thus, we introduce a conditional BiFPN-like diffusion architecture, which we term as the conditional-Mini-BiFPN (cMini-BiFPN). Note that we use the term cMini-BiFPN to indicate the bi-directional feature fusion alone, without the GSM module. As depicted in \cref{fig:Diffuser}, the initial three blocks $F^1_{in}, F^2_{in}, F^3_{in}$—the encoder blocks—process both partially masked original $\tilde{x}^F_0$ and perturbed BEV features $x^F_t$, where $\tilde{x}^F_0$ serves as a guide for the diffusion process, aiding in the progressive refinement of the corrupted BEV feature. Subsequent blocks $F^1_{out}, F^2_{out}, F^3_{out}$—the decoder blocks—accept the conditioned BEV representation and apply iterative top-down and bottom-up bi-directional fusion to generate the final BEV feature. Particularly, 
\begin{equation}
    \scalebox{0.9}{%
        $F_{it}^{2}=\operatorname{Conv}\left(\delta\left(\frac{\left(w_{1}^{\prime} \cdot F_{i n}^{2}+w_{2}^{\prime} \cdot Resize\left(F_{i n}^{3}\right)\right.}{w_{1}^{\prime}+w_{2}^{\prime}+\varepsilon}\right)\right)$}
    \label{eq:f2_up}
\end{equation}
\begin{equation}
    \scalebox{0.9}{%
        $F_{out}^{2}=\operatorname{Conv}\left(\delta\left(\frac{\left(w_{1}^{\prime\prime} \cdot F_{in}^{2} +w_{2}^{\prime\prime} \cdot F_{it}^{2}+ w_{3}^{\prime\prime} \cdot Resize\left(F_{out}^{1}\right) \right.}{w_{1}^{\prime\prime}+w_{2}^{\prime\prime}+w_{3}^{\prime\prime} +\varepsilon}\right)\right)$}
    \label{eq:f2}
\end{equation}
where $\delta$ denotes the Swish activation function \cite{efficientdet}, $F^2_{it}$ represents the intermediate fusion result of $F^2_{in}$ and $F^3_{in}$, and $F_{out}^{2}$ is computed by blending $F^2_{it}$, $F_{out}^{1}$, and $F_{in}^{2}$. The weights $w_{1,2,3}^{\prime}$ and $w_{1,2,3}^{\prime\prime}$ are learnable and optimised during training. $F_{out}^{1}$ and $F_{out}^{3}$ are derived similarly to $F_{out}^{2}$. The final feature map is the concatenation of $F_{out}^{1}$, $F_{out}^{2}$, and $F_{out}^{3}$, which is then processed through a Transfusion-L \cite{transfusion} detection head and a segmentation head \cite{bevfusionmit}, facilitating both 3D object detection and BEV map segmentation. 
We believe that this streamlined and efficient cMini-BiFPN architecture is well-suited for the demanding diffusion processes in autonomous driving applications, balancing computational efficiency with the necessity for expeditious sampling.

\subsection{Progressive Sensor Dropout Training (PSDT)}
Sensor reliability in autonomous driving is crucial. Cameras and LiDAR systems, the eyes of robotic agents, can occasionally be occluded or fail, potentially compromising safety and operational efficiency. Recognising this, we propose the Progressive Sensor Dropout Training (PSDT) method, which enhances our DifFUSER model's robustness and adaptability, particularly in scenarios where sensor inputs may be occluded or failed. It enables the model to generate or reconstruct missing features from camera or LiDAR inputs by leveraging the learned distribution across both modalities. This approach ensures consistently high performance, even in cases of sensor malfunction or when sensor data is partially missing, closely simulating real-world operational challenges.

Specifically in PSDT, $x^F_0$, the input for the whole fusion process, containing features from both camera and LiDAR, is utilised in three key ways: as a training target, as a noisy input for the diffusion module, and as a diffusion condition with randomly dropped camera or LiDAR features. To simulate the conditions of sensor dropout or malfunction, we progressively increase the dropout rate for either the camera or LiDAR inputs from 0\% to a predefined maximum $\alpha=25$ during training. This is represented as:
\begin{equation}
    F_{\text{masked}}^m = F^m \odot \text{Bernoulli}\left(\frac{\alpha}{100} \cdot \frac{e}{E}\right)
\end{equation}
where \(e\) represents the current epoch out of the total \(E\) epochs, defining the dropout probability \(p_{\text{dropout}}(e)\) that dictates the chance of each feature in modality \(F^m\) being dropped. This strategic process, achieved by applying a binary mask generated from a Bernoulli distribution based on \(p_{\text{dropout}}(e)\), not only trains the model to effectively denoise and generate more expressive features but also minimises its reliance on any single sensor, thereby enhance its capability to handle incomplete sensor data with greater resilience. 

\subsection{Gated Self-Conditioned Modulated (GSM) Diffusion Module}
Departing from traditional diffusion models \cite{ddpm, ddim}, which primarily use just $t$ as the condition, the GSM module leverages both the temporal (time step \(t\)) and the sensor information contained within the partially masked latent sample (\(\tilde{x}^F_0\)) to guide the diffusion process. This requires stronger conditioning on the noisy latent to shape the diffusion trajectory towards generating high-quality features. This dual-conditioning approach, augmented by the PSDT paradigm and task-specific losses, ensures the diffusion process focuses on feature enhancement and not mere replication, encourage the synthesis of more expressive features. 

\begin{wrapfigure}{r}{0.6\textwidth} 
  \centering
  \vspace{-2em}
  \includegraphics[width=0.58\textwidth]{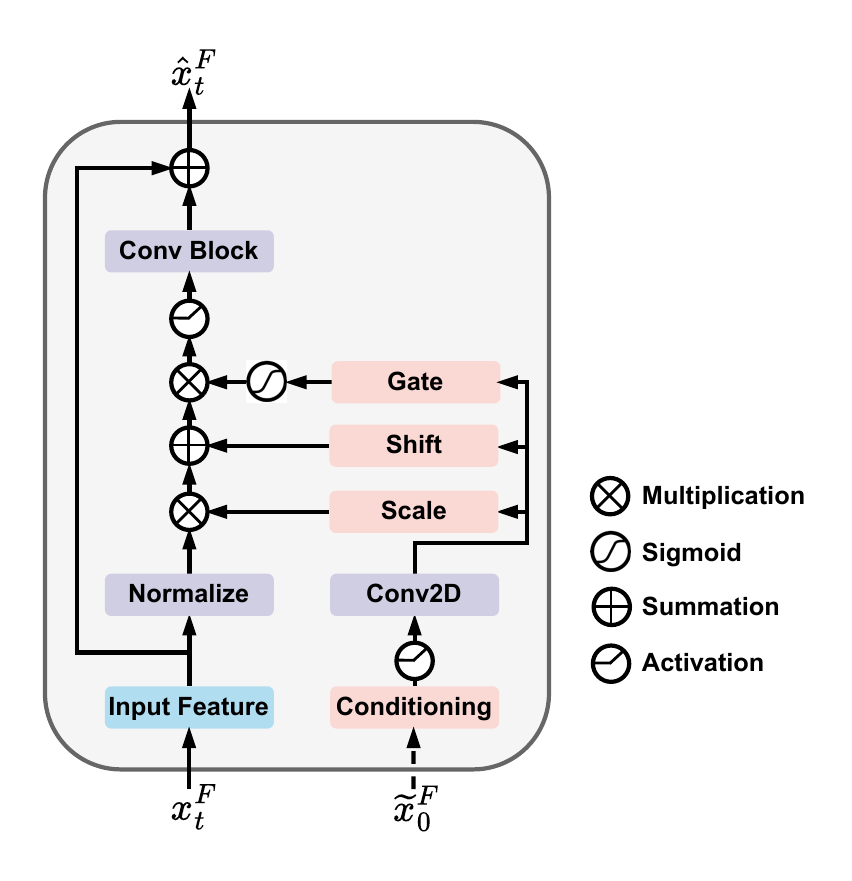} 
  \caption{GSM diffusion block.}
  \label{fig:diffuserblock}
  \vspace{-7em}
\end{wrapfigure}

Specifically, in DifFUSER, the GSM module is used in encoder blocks $F^1_{in}, F^2_{in}, F^3_{in}$ (see \cref{fig:Diffuser}). The GSM diffusion process is encapsulated by a FiLM-like (Feature-wise Linear Modulation) operation. This modulation computes parameters that dynamically adjust the diffusion trajectory based on the conditions $\tilde{x}^F_0$. The modulation formula is defined as:
\begin{equation}
    \begin{aligned}
        \tilde{x}^F_0 &= \tilde{x}^F_0 + \textit{time\_emb}(t), \\
        \gamma_t &= \sigma(W_{\gamma} \cdot \tilde{x}^F_0 + b_{\gamma}), \\
        \alpha_t &= W_{\alpha} \cdot \tilde{x}^F_0 + b_{\alpha}, \\
        \beta_t &= W_{\beta} \cdot \tilde{x}^F_0 + b_{\beta}, \\
        \hat{x}^F_0 &= \gamma_t \cdot (x^F_t \cdot (1+\alpha_t) + \beta_t).
    \end{aligned}
    \label{eq:GSM}
\end{equation}

Here, \(\tilde{x}^F_0\) is the version of \(x^F_0\) that may be partially masked due to PSDT. The parameters \(\alpha_t\), \(\beta_t\), and \(\gamma_t\) modulate the noisy input feature \(x^F_t\) via scaling and shifting, with \(\gamma_t\) acting as a sigmoid-gated self-conditioning element to recalibrate layer activation based on the feature itself. These parameters are derived from the condition \(\tilde{x}^F_0\) using a simple convolution block, we deliberately limit gradient flow when introducing noise to \(x^F_0\) to prevent model memorisation. The architecture of our gated self-conditioned modulated diffusion mechanism is detailed in \cref{fig:diffuserblock}.

\subsection{DifFUSER training and loss}
DifFUSER is trained end-to-end in either single-task or multi-task fashion, together with task-specific losses, which encourage the module to not just denoising but also generating more expressive BEV features serving the downstream tasks. In the forward noising process, Gaussian noise $\varepsilon$ is gradually added to the extracted features $x^F_0$ via a discrete \textit{stochastic differential equation} (SDE):
\begin{equation}
q(x^F_t, x^F_0) = \sqrt{\bar{\alpha_t}}x^F_0 + \varepsilon \sqrt{1 - \bar{\alpha_t}}
\end{equation}
where $\bar{\alpha_t}$ is a hyperparameter and $\varepsilon \sim \mathcal{N}(0, I)$. If the time step $t$ is large, $x^F_t$ would become Gaussian noise (see Appendix for more discussion).

The reverse denoising process is a discrete SDE that gradually maps a Gaussian noise into a sample. At each time step, given $x^F_t$, $t$, and $\tilde{x}^F_0$, it predicts the next reverse step:
\begin{equation}
    p^{\left( t \right)}(x^F_{t-1} | x^F_t, \tilde{x}^F_0, t) =
    \begin{cases}
        \mathcal{N}\left( p^{\left( 1 \right)}_\theta(x^F_1, \tilde{x}^F_0, t), \sigma^2I \right) & \text{if } t = 1 \\
        q(x^F_{t-1} | x^F_t, p^{\left( t \right)}_\theta(x^F_t,\tilde{x}^F_0, t)) & \text{otherwise.}
    \end{cases}
\end{equation}
DDPM sampler \cite{ddpm} is not considered a practical model due to its slow sampling speed. Later, we explore and compare the potential of using training-free faster diffusion samplers, such as the first-order sampler DDIM \cite{ddim}, or higher-order samplers like DPM-Solver++ \cite{dpmsolver++} and DEIS \cite{deis}.

\noindent\textbf{Diffusion loss}. Instead of using $\varepsilon$-prediction from DDPM \cite{ddpm}, we train our model to predict the denoised and unmasked sample $x^F_0$ from the noisy $x^F_t$ at each time step, and use a simple mean-squared error loss on the prediction, thanks to the parameterisation of $\mu_\theta$ as a denoising network $p_\theta$ :
\begin{equation}
    \label{eq:diffusion_loss}
    \mathcal{L}_{diffusion} = \mathbb{E}_{t \sim [1,T], \widetilde{x}^F_t \sim q_t} \left[ \left\| p^{\left( t \right)}_\theta(x^F_t,t, \tilde{x}^F_0) - x^F_0 \right\|^2 \right]
\end{equation}

where $\widetilde{x}^F_0$ is the original partially masked feature from both sensors, $x^F_t$ is the noisy feature at time step $t$, $q_t$ is the distribution of $x^F_t$ at time step $t$, and $p^{\left( t \right)}_\theta(x^F_t,t, x^F_0))$ is the predicted $\hat{x}^F_0$ at time step $t$.

\noindent\textbf{Segmentation loss}. We use focal loss $\mathcal{L}_{seg}$ for the segmentation head, as in \cite{bevfusionmit}.

\noindent\textbf{Detection loss}. Similar to \cite{bevfusionmit, transfusion}, the Hungarian set prediction loss is applied to the detection head, which is a combination of the classification focal loss and the smooth L1 regression loss. Top $K$ predictions are selected and matched with the ground truth boxes by Hungarian matching. The detection loss is then calculated as 
$\mathcal{L}_{detection} = \lambda_{cls} \mathcal{L}_{cls} + \lambda_{reg} \mathcal{L}_{reg}$.

\noindent\textbf{Total loss}. The total loss is the weighted sum of the diffusion loss, the detection loss, and the segmentation loss, depending on the training mode:
\begin{equation}
\label{eq:total_loss}
\mathcal{L}_{total} = \mathcal{L}_{diffusion} + \lambda_{seg} \mathcal{L}_{seg} + \lambda_{det} \mathcal{L}_{detection}
\end{equation}
where $\lambda_{diff}, \lambda_{det}, \lambda_{seg}$ are the weights for each loss.

\section{Experiment}
\label{sec:experiments}
Our experiments focus on two critical tasks in autonomous driving: 3D object detection and Bird-Eye-View (BEV) map segmentation. 3D object detection is essential for accurately identifying and locating objects within a 3D space, such as vehicles, pedestrians, and road obstacles. BEV map segmentation, on the other hand, involves partitioning the bird's-eye view of the driving environment into different semantic categories, aiding in path planning and navigation. Both tasks benefit significantly from the fusion of LiDAR and camera data, offering comprehensive spatial and contextual understanding.

\noindent\textbf{Dataset.} We report DifFUSER's performance on the nuScenes \cite{nuscenes} dataset, a challenging large-scale benchmark with multiple multi-modal data annotations for autonomous driving (3D bounding box and map segmentation), which is suitable for our method. This dataset comprises roughly 1000 scenes, split into 700 for training, 150 for validation, and another 150 for testing. Each scene in the dataset represents a 20-second duration, recorded using six cameras and a LiDAR, along with additional data from five radars. The cameras, calibrated for precision, capture RGB images at a rate of 12 FPS, encompassing a full 360-degree horizontal field of view. Simultaneously, the 32-beam LiDAR scans the environment at 20 FPS. We do not use Radar data in our experiments.

\noindent\textbf{BEV map segmentation.} BEV map segmentation, an integral part of autonomous driving, involves segmenting the bird's-eye view of the road into distinct semantic categories. This task aids in path planning and navigation by providing a comprehensive overview of the road layout and relevant features. We adopt the training approach of the baseline BEVFusion \cite{bevfusionmit}, assessing IOUs of \textit{drivable areas, pedestrian crossings, walkways, stop lines, carparks, dividers}, and class-averaged mean IoU (mIOU). Our evaluation is conducted in a 100m x 100m area centered around the ego vehicle and reports mIOU.

\noindent\textbf{3D object detection.} Next, we evaluate DifFUSER on 3D object detection. NuScenes \cite{nuscenes} dataset provides 3D bounding boxes for 10 object classes, namely \textit{car, truck, bus, trailer, construction vehicle, pedestrian, motorcycle, bicycle, traffic cone, barrier}. Adhering to the dataset's established protocols, we transform the data from the preceding nine frames into the current point cloud frame for both training and evaluation. We use the official evaluation metrics for 3D object detection, namely \textit{NDS} and \textit{mAP}. And mean Average Translation Error (mATE), mean Average Scale Error (mASE), mean Average Orientation Error (mAOE), mean Average Velocity Error (mAVE), and mean Average Attribute Error (mAAE) are reported. Higher \textit{NDS} and \textit{mAP} indicate better performance.

\noindent\textbf{Training Details.} Unless specified otherwise, all models are trained using PSDT, with a batch size of 3 on two NVIDIA 4090 GPUs. Following \cite{bevfusionmit}, we keep Swin-T \cite{swintransformer} and Voxelnet \cite{voxelnet} as image and LiDAR backbones, respectively. We reduce the resolution of camera images to 448×800 pixels and convert the LiDAR point cloud into voxels, using a size of 0.075 meters for detection and 0.1 meters for segmentation. To accommodate different spatial requirements of detection and segmentation, we use grid sampling with bilinear interpolation to adjust BEV feature maps before each specific task. For 3D object detection, we first pretrain LiDAR only backbone for 15 epochs, then finetune the fusion module for 10 epochs with AdamW \cite{adamw} optimizer with a learning rate of 3.75e-5 and a Cosine Annealing scheduler \cite{cosineannealing}. For BEV Map Segmentation, we train the model for 20 epochs with AdamW optimizer with a learning rate of 2e-5 and a OneCycle scheduler \cite{onecycle}. We use the same training strategy in our ablation study for fair comparison. We also study joint 3D detection and segmentation training in our ablation study using a shorter training schedule for both tasks.

\section{Dicussion and Comparison with State-of-the-arts}

\begin{table*}[t]
    \centering
    \caption{DifFUSER outperforms state-of-the-art multi-sensor fusion methods on BEV map segmentation on nuScenes \textit{val} with a 6.4\% improvement across categories. Abbreviations: Mod. = Modality, Driv. = Drivable, Ped. Cr. = Pedestrian Crossing, Wlk. = Walkway, Stp. Ln. = Stop Line, Cpk. = Carpark, Div. = Divider. Relative improvements over the baseline are shown below. Bold, blue, and underlined values indicate the best, second-best, and baseline performance, respectively.}
    \begin{adjustbox}{width=\linewidth,center}
        \begin{tabular}{|l|c|cccccc|c|}
            \hline
            \textbf{Method} & \textbf{Mod.$\uparrow$} & \textbf{Driv.$\uparrow$} & \textbf{Ped. Cr.$\uparrow$} & \textbf{Wlk.$\uparrow$} & \textbf{Stp. Ln.$\uparrow$} & \textbf{Cpk.$\uparrow$} & \textbf{Div.$\uparrow$} & \textbf{Mean$\uparrow$} \\
            \hline
            \hline
            LSS \cite{lss}         & C            & 75.4  & 38.8  & 46.3  & 30.3  & 39.1  & 36.5  & 44.4 \\
            CVT \cite{cvt}            & C            & 74.3  & 36.8  & 39.9  & 25.8  & 35.0  & 29.4  & 40.2 \\
            M$^2$BEV \cite{m2bev}       & C            & 77.2  & -     & -     & -     & -     & 40.5  & -    \\
            BEVFusion \cite{bevfusionmit}       & C            & 81.7  & 54.8  & 58.4  & 47.4  & 50.7  & 46.4  & 56.6 \\
            \hline
            PointPillars \cite{pointpillars}    & L            & 72.0  & 43.1  & 53.1  & 29.7  & 27.7  & 37.5  & 43.8 \\
            CenterPoint \cite{centerpoint}     & L            & 75.6  & 48.4  & 57.5  & 36.5  & 31.7  & 41.9  & 48.6 \\
            \hline
            PointPainting \cite{pointpainting}  & LC           & 75.9  & 48.5  & 57.1  & 36.9  & 34.5  & 41.9  & 49.1 \\
            MVP            & LC           & 76.1  & 48.7  & 57.0  & 36.9  & 33.0  & 42.2  & 49.0 \\
            \underline{BEVFusion} \cite{bevfusionmit}& LC & \underline{85.5} & \underline{60.5} & \underline{67.6} & \underline{52.0} & \underline{57.0} & \underline{53.7} & \underline{62.7} \\
            \hline
            \hline
            \textbf{DifFUSER (Ours)} & LC & \textbf{89.7} & \textbf{69.0} & \textbf{74.3} & \textbf{61.9} & \textbf{63.0} & \textbf{62.3} & \textbf{70.0} \\
            Improvement & & $\uparrow$4.2 & $\uparrow$8.5 & $\uparrow$6.7 & $\uparrow$9.9 & $\uparrow$6.0 & $\uparrow$8.6 & $\uparrow$7.3 \\
            \hline
        \end{tabular}
    \end{adjustbox}
    \label{tab:val_seg_performance}
\end{table*}

\begin{figure*}[tp]
    \centering
    \includegraphics[width=\linewidth]{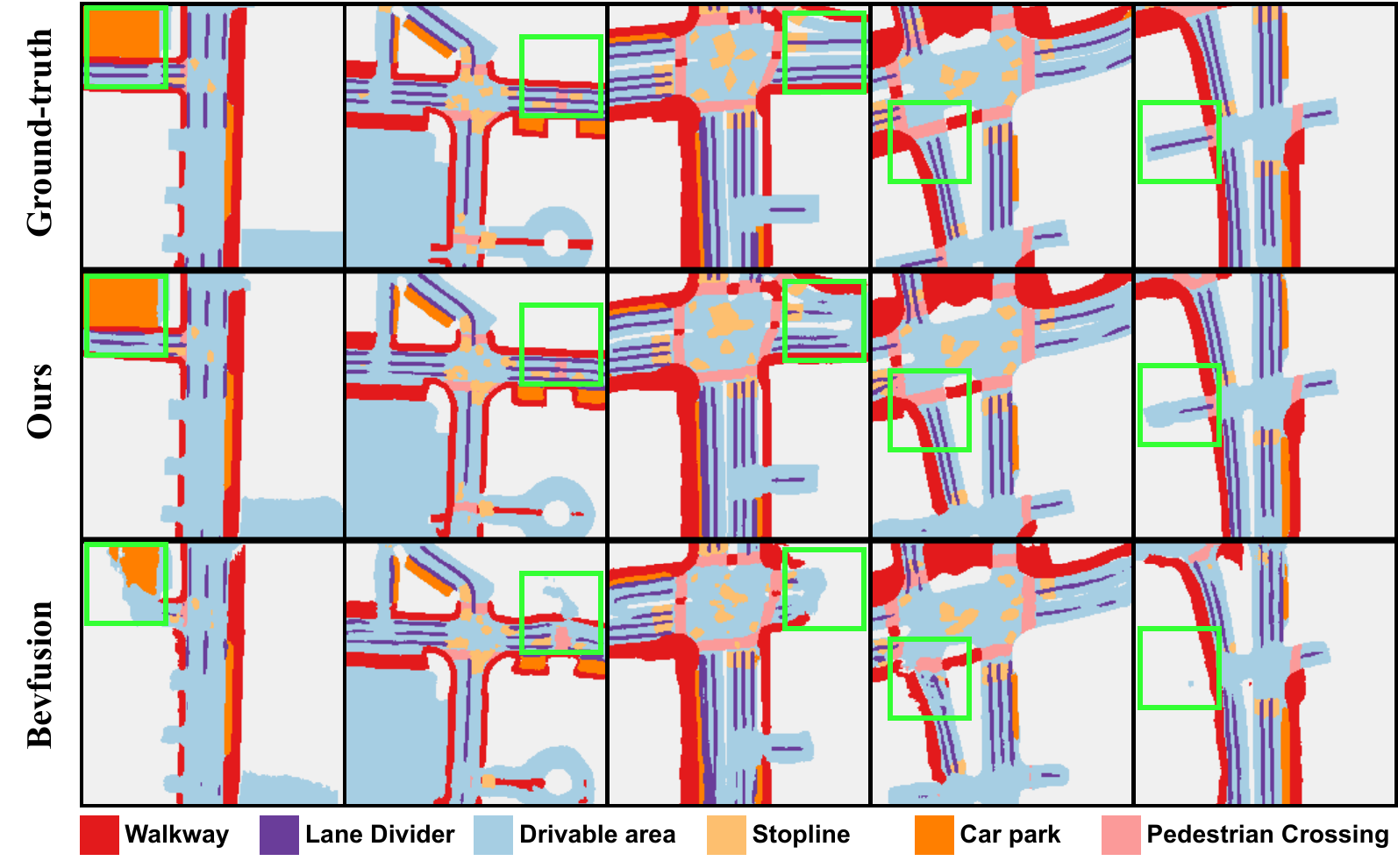}
    \caption{Qualitative result on BEV map segmentation task on the Nuscenes \textit{val} set, compared to BEVFusion \cite{bevfusionmit}. Green bounding boxes highlight the areas where the baseline method fails to segment, while our method can segment correctly.}
    \vspace{-1.5em}
    \label{fig:seg_comparison}
\end{figure*}

\noindent\textbf{BEV Map Segmentation.} \cref{tab:val_seg_performance} shows the comparison of BEV map segmentation performance on the NuScenes \textit{val} set. DifFUSER exhibits a remarkable performance leap over the baseline BEVFusion \cite{bevfusionmit}. While BEVFusion \cite{bevfusionmit} achieves an mIOU score of 62.7, DifFUSER significantly elevates this to 70.04, marking a substantial 7.3$\%$ improvement. This pronounced advancement is consistent across various categories, as in \cref{tab:val_seg_performance}
all outperforming their respective scores in BEVFusion \cite{bevfusionmit}. These results underscore DifFUSER's capability to enhance the quality of fused features, which is crucial in the segmentation task. \cref{fig:seg_comparison} shows the qualitative results of DifFUSER's BEV map segmentation on the NuScenes \textit{val} set. BEVFusion tends to produce noisy segmentation results, especially at further distances, where the LiDAR point cloud is sparse and sensor misalignment is more pronounced. In contrast, DifFUSER produces more accurate segmentation results, with finer details and less noise.

\begin{table*}[!t]
   \centering
    \caption{Comparison of 3D object detection performance on the Nuscenes \textit{test} dataset. Mod. abbreviates Modality, where L and C denote LiDAR and Camera, respectively. Relative improvement over the baseline is shown in a separate row. Bold, blue, and underlined values indicate the best, second-best, and baseline performance, respectively.}
   \begin{adjustbox}{width=\linewidth,center}
   \begin{tabular}{|l|c|cc|ccccc|}
       \hline
       \textbf{Methods} & \textbf{Mod.$\uparrow$} & \textbf{NDS$\uparrow$}  & \textbf{mAP$\uparrow$}  & \textbf{mATE$\downarrow$} & \textbf{mASE$\downarrow$} & \textbf{mAOE$\downarrow$} & \textbf{mAVE$\downarrow$} & \textbf{mAAE$\downarrow$} \\
       \hline
       \hline
       PointPainting \cite{pointpainting} & LC & 61.0 & 54.1 & 0.380 & 0.260 & 0.541 & 0.293 & 0.131 \\
       PointAugmenting \cite{pointaugmenting} & LC & 71.1 & 66.8 & 0.253 & 0.235 & 0.354 & 0.266 & \textcolor{blue}{0.123} \\
       MVP \cite{mvp} & LC & 70.5 & 66.4 & 0.263 & 0.238 & 0.321 & 0.313 & 0.134 \\
       FusionPainting \cite{fusionpainting} & LC & 71.6 & 68.1 & 0.256 & 0.236 & 0.346 & 0.274 & 0.132 \\
       UVTR \cite{uvtr} & LC & 71.1 & 67.1 & 0.306 & 0.245 & 0.351 & \textcolor{blue}{0.225} & 0.124 \\
       TransFusion \cite{transfusion} & LC & 71.7 & 68.9 & 0.259 & 0.243 & 0.359 & 0.288 & 0.127 \\
       BEVFusion \cite{bevfusionpeking} & LC & 71.8 & 69.2 & 0.260 & 0.250 & 0.370 & 0.270 & 0.130 \\
       \underline{BEVFusion} \cite{bevfusionmit} & LC & \underline{72.9} & \underline{70.2} & \underline{0.261} & \underline{0.239} & \underline{0.329} & \underline{0.260} & \underline{0.134} \\
       DeepInteraction \cite{deepinteraction} & LC & 73.4 & 70.8 & 0.257 & 0.240 & 0.325 & 0.245 & 0.128 \\
       MSMDFusion \cite{msmdfusion} & LC & 73.0 & 71.0 &\textcolor{blue}{0.250} & 0.240 & 0.330 & 0.260 & 0.140 \\
       LinK \cite{link} & LC & 73.4 & 69.8 & 0.258 & \textcolor{blue}{0.229} & 0.312 & 0.234 & 0.136 \\
       FocalFormer3D \cite{focalformer3d} & LC & \textcolor{blue}{73.9} & \textcolor{blue}{71.6} & \textcolor{blue}{0.250} & 0.240 & 0.330 & \textbf{0.220} & 0.130 \\
       LargeKernel-F \cite{largekernel3d} & LC & \textbf{74.1} & 71.1 & \textbf{0.236} & \textbf{0.228} & \textbf{0.298} & 0.241 & 0.131 \\
      CMT \cite{cmt} & LC & \textbf{74.1} & \textbf{72.0} & 0.279 & 0.235 & \textcolor{blue}{0.308} & 0.259 & \textbf{0.112} \\

       \hline
       \hline
       \textbf{DifFUSER (ours)} & LC & 73.8 & 71.3 & \textcolor{blue}{0.250} & 0.241 & 0.317 & 0.244 & 0.130 \\
       Improvement & & $\uparrow$1.9 & $\uparrow$1.0 & $\downarrow$0.011 & $\uparrow$0.002 & $\downarrow$0.012 & $\downarrow$0.016 & $\downarrow$0.004\\
       \hline
   \end{tabular}
   \end{adjustbox}
   \label{tab:test_det_performance}
\end{table*}

\begin{figure}[!t]
    \centering
    \begin{minipage}[t]{.49\textwidth} 
        \centering
        \vspace{-1.1em}
        \captionof{table}{Results for 3D object detection on NuScenes (val). Bold, blue, and underlined values indicate the best, second-best, and baseline performance, respectively. Mod. stands for Modality.}
        \vspace{0.2em}
        \adjustbox{width=\linewidth}{ 
        \begin{tabular}{|l|c|c|c|}
            \hline
            \textbf{Methods} & \textbf{Mod.} & \textbf{NDS$\uparrow$} & \textbf{mAP$\uparrow$} \\
            \hline
            \hline
            TransFusion \cite{transfusion} & L & 70.1 & 65.1 \\
            FUTR3D \cite{futr3d} & LC & 68.3 & 64.5 \\
            UVTR \cite{uvtr} & LC & 70.2 & 65.4 \\
            TransFusion \cite{transfusion} & LC & 71.3 & 67.5 \\
            \underline{BEVFusion} \cite{bevfusionmit} & \underline{LC} & \underline{71.4} & \underline{68.5} \\
            DeepInteration \cite{deepinteraction} & LC & 72.6 & 69.9 \\
            CMT \cite{cmt} & LC & \textbf{72.9} & \textbf{70.3} \\
            \hline
            \hline
            \textbf{DifFUSER} & LC & \textcolor{blue}{72.6} & \textcolor{blue}{70.0} \\
            Improvement & & $\uparrow$1.2& $\uparrow$1.5 \\
            \hline
        \end{tabular}
        }
        \label{tab:val_det_performance}
    \end{minipage}%
    \hspace{0.1em}
    \begin{minipage}[t]{.49\textwidth} 
        \centering
        \captionof{figure}{Performance Comparison of DifFUSER with and without Sensor Drop (either Lidar or Camera)}
        \vspace{-0.2em} 
        \includegraphics[width=1.125\linewidth]{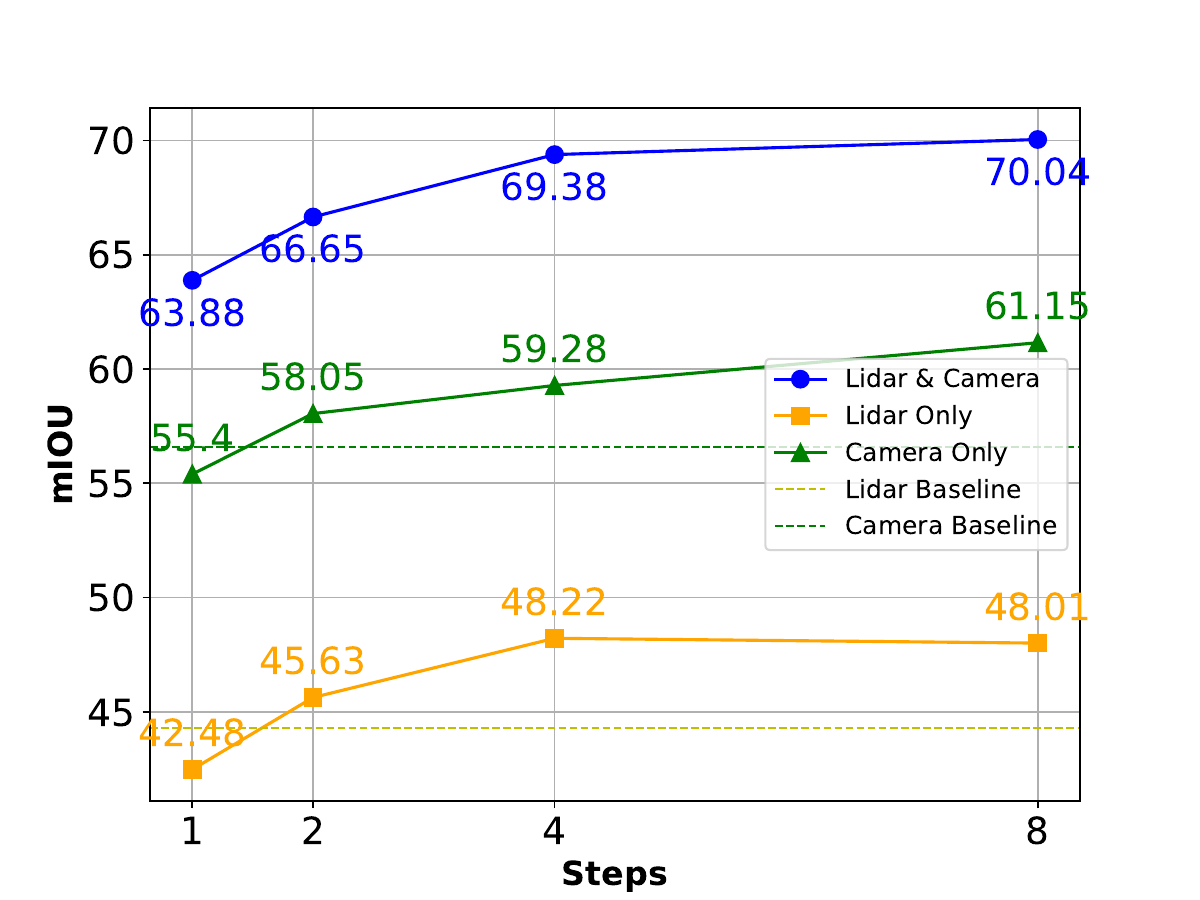}
        \label{fig:sensor_drop}
    \end{minipage}
    \vspace{-1.5em}
\end{figure}

\noindent\textbf{3D Object Detection.} \cref{tab:test_det_performance} shows the comparison of 3D object detection performance on the NuScenes \textit{test} set. DifFUSER demonstrates notable improvement over the baseline BEVFusion \cite{bevfusionmit}. While BEVFusion \cite{bevfusionmit} already showed impressive results with an NDS of 72.9 and an mAP of 70.2, DifFUSER further enhances these metrics to 73.8 (+0.9$\%$) and 71.2 (+1$\%$), competing closely with CMT \cite{cmt}, which achieves an NDS of 74.1 and an mAP of 72.0. This improvement is attributed to the more effective multi-modal diffusion-based fusion. 
 Specifically, DifFUSER's denoising capability contributes to finer detail preservation and noise reduction in fused features, leading to better object detection accuracy. The improvement is also evident in other metrics, with DifFUSER achieving lower error rates (mATE, mASE, mAOE, mAVE, mAAE) compared to BEVFusion \cite{bevfusionmit}. This suggests that DifFUSER not only enhances the detection accuracy but also improves the overall reliability and precision of the object detection task. The same trend is observed in the validation set, as in \cref{tab:val_det_performance}.

\noindent\textbf{Robustness to sensor failure.} To highlight the robustness of DifFUSER under sensor failures, we present the segmentation performance comparison in \cref{fig:sensor_drop}. The findings reveal that with a sufficient number of sampling steps, the model is capable of generating features that effectively compensate for and act as replacements for the missing sensor data. This capability is a significant advancement over existing models like CMT\cite{cmt}, which, under similar missing sensor conditions, can only match the performance of single sensor baselines. Our DifFUSER model distinguishes itself by not just matching but exceeding the baseline performance achieved with single sensor inputs thanks to the synthetic features it generates. The ability to generate and utilise synthetic features effectively mitigates the reliance on any single sensor modality, ensuring the model's operational resilience in diverse and challenging environments.

\section{Ablation Study}
\label{ablation}

\noindent\textbf{GSM diffusion block design.} We conducted experiments about the components of the GSM module, where all models are trained with a shorter training schedule (6 epochs). As shown in \cref{tab:GSM}, by incrementally incorporating Scale, Shift, and Gate operations, we observed a marked improvement in mean Intersection over Union (mIOU), with the Gate operation significantly boosting the mIOU to 62.17. This underscores the effectiveness of GSM's dynamic feature modulation, especially its capacity for strong conditioning, which selectively emphasises relevant features, enabling the model to generate meaningful outputs.

\noindent\textbf{Contribution of each component.} We investigate the contribution of each component in DifFUSER to understand whether the improvement comes from the enhanced cMini-BiFPN fusion architecture or the gated self-conditioned latent diffusion module. We conduct experiments on the NuScenes \textit{val} set with the same settings as in \cref{sec:experiments}, using single-task models. It can be seen that by adding BiFPN-like feature fusion module, the performance of BEV map segmentation and 3D object detection are improved by 3.8\% in mIOU and 0.4\% in NDS, respectively. 
Further integrating the GSM module with a 4-step sampling improves mIOU by an additional 2.3\% and NDS by an additional 0.3\%. Extending the diffusion sampling to 8 steps further raises mIOU to 70.04\% and NDS to 72.6\%, which is the final performance of DifFUSER. 

\noindent\textbf{Diffusion ODE Solvers.} We perform a small experiment to compare the performance of other second-order ODE solvers besides the de-facto standard first-order DDIM solver. We choose BEV map segmentation for this experiment and report the segmentation mIOU results. We use the same settings as those in \cref{sec:experiments}, and compare the performance of DDIM \cite{ddim}, DPM-Solver++ \cite{dpmsolver++} and DEIS \cite{deis} at different sampling steps. As shown in \cref{tab:sampling_methods_performance}, the performance of DPM-Solver++ and DEIS are comparable to DDIM, and the performance of all three methods are improved as the number of sampling steps increases. It shows that all solvers can lead to good convergence of the fused features, and the performance is not sensitive to the choice of solvers. For best performance, We choose DDIM with 8 sampling steps as the default solver in our experiments due to its simplicity and efficiency. However, in terms of efficiency, we believe 4-step sampling is already sufficient.

\noindent\textbf{Inference Time.} Here we measure the inference time of DifFUSER on a single NVIDIA 4090 GPU. At 1-step sampling, Diffuser can reach an average speed of ~8 FPS (123.1 ms latency), the full break-down is shown in \cref{tab:inference_speed}
\begin{table*}[t]
    \begin{minipage}{.52\linewidth}
        \centering
        \captionof{table}{Performance comparison of different sampling methods at varying steps}
        \begin{adjustbox}{height=1cm, center}
        \begin{tabular}{|l|cccc|}
            \hline
            \textbf{Sampling step} & \textbf{1} & \textbf{2} & \textbf{4} & \textbf{8} \\
            \hline
            \hline
            DDIM\cite{ddim}                 & 63.88      & 66.65      & 68.82      & \textbf{70.04}       \\
            DPM-Solver\cite{dpmsolver++}          & 64.02      & 66.20      & 68.90      & \textbf{70.02}       \\
            DeIS\cite{deis}                  & 63.80      & 66.30      & 68.81      & \textbf{69.95}       \\
            \hline
        \end{tabular}
        \end{adjustbox}
        \label{tab:sampling_methods_performance}
    \end{minipage}%
    \hspace{0.2em}
    \begin{minipage}{.46\linewidth}
        \centering
        \captionof{table}{Multi-task joint training performance}
        \adjustbox{height=1cm}{
            
        \begin{tabular}{|l|c|c|}
            \hline
            \textbf{Task} & \textbf{NDS} & \textbf{mIOU} \\
            \hline
            \hline
            Detection   & 69.59 & -     \\
            Segmentation   & -     & 65.8  \\
            Joint  & 68.23 & 61.36 \\
            \hline
        \end{tabular}
        }
        \label{tab:multitask_performance}
    \end{minipage}
\end{table*}
\begin{table*}[t]
    \centering
    \begin{minipage}{.52\linewidth}
        \vspace{-0pt} 
        \centering
        \captionof{table}{Comparison of Detection and Segmentation Performance}
        \vspace{-1em} 
        \begin{adjustbox}{height=1.06cm, center}
        \begin{tabular}{|l|cc|c|}
            \hline
            \textbf{Component} & \textbf{mAP} & \textbf{NDS} & \textbf{mIOU} \\
            \hline
            \hline
            Baseline**        & 68.9         & 71.4         & 62.7          \\
            +cMiniBiFPN**       & 69.4        & 71.8         & 66.5          \\
            +GSM (4-step) + PSDT* & 69.7      & 72.1         & 68.82          \\
            +GSM (8-step) + PSDT* & \textbf{70.0}       & \textbf{72.6}         & \textbf{70.04}         \\
            \hline
        \end{tabular}
        \end{adjustbox}
        \scriptsize{[*] With $\mathcal{L}_{diffusion}$, [**] Without $\mathcal{L}_{diffusion}$}
    \end{minipage}%
    \hspace{0.2em}
    \begin{minipage}{.46\linewidth}
        \vspace{-1.8em}
        \centering
        \caption{GSM diffusion block design.}
        \vspace{0.1em}
        \adjustbox{height=0.95cm}{
        \begin{tabular}{|ccc|c|}
            \hline
            \textbf{Scale} & \textbf{Shift} & \textbf{Gate} & \textbf{mIOU} \\
            \hline
            \hline
            \cmark & & & 59.92\\
            \cmark & \cmark & & 60.73 \\
            \cmark & \cmark & \cmark & \textbf{62.17} \\
            \hline
        \end{tabular}
        }
        \label{tab:GSM}
    \end{minipage}
\end{table*}

\begin{table}[!t]\centering
\caption{Detailed breakdown of inference speeds, excluding the speed of point cloud voxelisation module, following our baseline.}
\begin{adjustbox}{width=0.7\columnwidth,center}
\begin{tabular}{|l|cccc|}\hline
\textbf{Module} & \textbf{Latency} & \textbf{Parameters} & \textbf{GFLOPs} & \textbf{MACs} \\\hline \hline
Camera  & 63ms\footnotemark & 27.5M & 127.7 & 63.8 \\
LiDAR  & 25ms & 7.1M & 32.3 & 16.2 \\
Fuser & 24ms & 37.5M & 294.0 & 147.0 \\
Task heads & 11ms & 4.7M & 189.0 & 94.5 \\
\textbf{Total} & 123ms & 76.8M & 643.0 & 321.5 \\
\hline
\end{tabular}
\end{adjustbox}
\label{tab:inference_speed}
\end{table}
\footnotetext{Including BEV-Pooling, a non-DL operation without extra parameters.}

\noindent\textbf{Multi-task joint training.} Finally, we investigate the performance of DifFUSER in multi-task joint training. We use a shorter training schedule and re-scale the losses of each task to balance the training. As shown in \cref{tab:multitask_performance}, the performance of both tasks decreases compared to single-task training, an issue commonly known as negative transfer, which we leave for future investigation. 

\section{Conclusion}
In this work, we propose DifFUSER, a diffusion-based generative model for multi-modal fusion in 3D perception. DifFUSER is designed to enhance the quality of fused features by improving the fusion architecture and leveraging the denoising property of diffusion models. Central to our approach is the introduction of a Gated Self-Conditioned Modulated (GSM) latent diffusion module and PSDT paradigm, specifically engineered to improve the fusion output features and robustness against sensor noise and failure. Additionally, cMini-BiFPN fusion architecture emerged as a promising multi-scale latent diffusion module. Extensive experiments on the NuScenes dataset show that DifFUSER achieves SOTA performance in BEV map segmentation tasks and can fairly compete with SOTA transformer-based fusion methods in 3D object detection. This shows the potential of generative models to significantly enhance the capability of autonomous driving systems. Our approach opens a direction of using diffusion models for better accuracy and reliability in 3D perception tasks, promising further enhancements and broader applicability of diffusion-based models in complex, real-world environments.

\noindent{\textbf{Acknowledgments.} This work has been partially funded by The Australian Research Council Discovery Project (ARC DP2020102427). We acknowledge the partial sponsorship of our research by DARPA Assured Neuro Symbolic Learning and Reasoning (ANSR) program, under award number FA8750-23-2-1016.}

%
%
\bibliographystyle{splncs04}
\bibliography{main}

\begin{thebibliography}{10}
\providecommand{\url}[1]{\texttt{#1}}
\providecommand{\urlprefix}{URL }
\providecommand{\doi}[1]{https://doi.org/#1}

\bibitem{transfusion}
Bai, X., Hu, Z., Zhu, X., Huang, Q., Chen, Y., Fu, H., Tai, C.L.: Transfusion: Robust lidar-camera fusion for 3d object detection with transformers. In: CVPR. pp. 1090--1099 (2022)

\bibitem{nuscenes}
Caesar, H., Bankiti, V., Lang, A.H., Vora, S., Liong, V.E., Xu, Q., Krishnan, A., Pan, Y., Baldan, G., Beijbom, O.: nuscenes: A multimodal dataset for autonomous driving. In: CVPR. pp. 11621--11631 (2020)

\bibitem{diffusiondet}
Chen, S., Sun, P., Song, Y., Luo, P.: Diffusiondet: Diffusion model for object detection. In: ICCV. pp. 19830--19843 (2023)

\bibitem{futr3d}
Chen, X., Zhang, T., Wang, Y., Wang, Y., Zhao, H.: Futr3d: A unified sensor fusion framework for 3d detection. In: CVPR. pp. 172--181 (2023)

\bibitem{focalformer3d}
Chen, Y., Yu, Z., Chen, Y., Lan, S., Anandkumar, A., Jia, J., Alvarez, J.M.: Focalformer3d: focusing on hard instance for 3d object detection. In: Proceedings of the IEEE/CVF International Conference on Computer Vision. pp. 8394--8405 (2023)

\bibitem{focalconv}
Chen, Y., Li, Y., Zhang, X., Sun, J., Jia, J.: Focal sparse convolutional networks for 3d object detection. In: CVPR. pp. 5428--5437 (2022)

\bibitem{largekernel3d}
Chen, Y., Liu, J., Zhang, X., Qi, X., Jia, J.: Largekernel3d: Scaling up kernels in 3d sparse cnns. In: CVPR. pp. 13488--13498 (2023)

\bibitem{rangedet}
Fan, L., Xiong, X., Wang, F., Wang, N., Zhang, Z.: Rangedet: In defense of range view for lidar-based 3d object detection. In: ICCV. pp. 2918--2927 (2021)

\bibitem{ddpm}
Ho, J., Jain, A., Abbeel, P.: Denoising diffusion probabilistic models. NeurIPS  \textbf{33},  6840--6851 (2020)

\bibitem{cfg}
Ho, J., Salimans, T.: Classifier-free diffusion guidance. arXiv preprint arXiv:2207.12598  (2022)

\bibitem{motiondiffuser}
Jiang, C., Cornman, A., Park, C., Sapp, B., Zhou, Y., Anguelov, D., et~al.: Motiondiffuser: Controllable multi-agent motion prediction using diffusion. In: CVPR. pp. 9644--9653 (2023)

\bibitem{msmdfusion}
Jiao, Y., Jie, Z., Chen, S., Chen, J., Ma, L., Jiang, Y.G.: Msmdfusion: Fusing lidar and camera at multiple scales with multi-depth seeds for 3d object detection. In: Proceedings of the IEEE/CVF conference on computer vision and pattern recognition. pp. 21643--21652 (2023)

\bibitem{pointpillars}
Lang, A.H., Vora, S., Caesar, H., Zhou, L., Yang, J., Beijbom, O.: Pointpillars: Fast encoders for object detection from point clouds. In: CVPR. pp. 12697--12705 (2019)

\bibitem{pifenet}
Le, D.T., Shi, H., Rezatofighi, H., Cai, J.: Accurate and real-time 3d pedestrian detection using an efficient attentive pillar network. RA-L  \textbf{8}(2),  1159--1166 (2022)

\bibitem{uvtr}
Li, Y., Chen, Y., Qi, X., Li, Z., Sun, J., Jia, J.: Unifying voxel-based representation with transformer for 3d object detection. NeurIPS  \textbf{35},  18442--18455 (2022)

\bibitem{bevdepth}
Li, Y., Ge, Z., Yu, G., Yang, J., Wang, Z., Shi, Y., Sun, J., Li, Z.: Bevdepth: Acquisition of reliable depth for multi-view 3d object detection. In: AAAI. vol.~37, pp. 1477--1485 (2023)

\bibitem{bevformer}
Li, Z., Wang, W., Li, H., Xie, E., Sima, C., Lu, T., Qiao, Y., Dai, J.: Bevformer: Learning bird’s-eye-view representation from multi-camera images via spatiotemporal transformers. In: European conference on computer vision. pp. 1--18. Springer (2022)

\bibitem{bevfusionpeking}
Liang, T., Xie, H., Yu, K., Xia, Z., Lin, Z., Wang, Y., Tang, T., Wang, B., Tang, Z.: Bevfusion: A simple and robust lidar-camera fusion framework. NeurIPS  \textbf{35},  10421--10434 (2022)

\bibitem{magic3d}
Lin, C.H., Gao, J., Tang, L., Takikawa, T., Zeng, X., Huang, X., Kreis, K., Fidler, S., Liu, M.Y., Lin, T.Y.: Magic3d: High-resolution text-to-3d content creation. In: CVPR. pp. 300--309 (2023)

\bibitem{zero123}
Liu, R., Wu, R., Van~Hoorick, B., Tokmakov, P., Zakharov, S., Vondrick, C.: Zero-1-to-3: Zero-shot one image to 3d object. In: ICCV. pp. 9298--9309 (2023)

\bibitem{swintransformer}
Liu, Z., Lin, Y., Cao, Y., Hu, H., Wei, Y., Zhang, Z., Lin, S., Guo, B.: Swin transformer: Hierarchical vision transformer using shifted windows. In: ICCV. pp. 10012--10022 (2021)

\bibitem{tanet}
Liu, Z., Zhao, X., Huang, T., Hu, R., Zhou, Y., Bai, X.: Tanet: Robust 3d object detection from point clouds with triple attention. In: AAAI. vol.~34, pp. 11677--11684 (2020)

\bibitem{bevfusionmit}
Liu, Z., Tang, H., Amini, A., Yang, X., Mao, H., Rus, D.L., Han, S.: Bevfusion: Multi-task multi-sensor fusion with unified bird's-eye view representation. In: 2023 IEEE International Conference on Robotics and Automation (ICRA). pp. 2774--2781. IEEE (2023)

\bibitem{cosineannealing}
Loshchilov, I., Hutter, F.: Sgdr: Stochastic gradient descent with warm restarts. arXiv preprint arXiv:1608.03983  (2016)

\bibitem{adamw}
Loshchilov, I., Hutter, F.: Decoupled weight decay regularization. arXiv preprint arXiv:1711.05101  (2017)

\bibitem{dpmsolver++}
Lu, C., Zhou, Y., Bao, F., Chen, J., Li, C., Zhu, J.: Dpm-solver++: Fast solver for guided sampling of diffusion probabilistic models. arXiv preprint arXiv:2211.01095  (2022)

\bibitem{link}
Lu, T., Ding, X., Liu, H., Wu, G., Wang, L.: Link: Linear kernel for lidar-based 3d perception. In: Proceedings of the IEEE/CVF Conference on Computer Vision and Pattern Recognition. pp. 1105--1115 (2023)

\bibitem{frustum-pointpillars}
Paigwar, A., Sierra-Gonzalez, D., Erkent, {\"O}., Laugier, C.: Frustum-pointpillars: A multi-stage approach for 3d object detection using rgb camera and lidar. In: ICCV. pp. 2926--2933 (2021)

\bibitem{clocs}
Pang, S., Morris, D., Radha, H.: Clocs: Camera-lidar object candidates fusion for 3d object detection. In: IROS. pp. 10386--10393. IEEE (2020)

\bibitem{fastclocs}
Pang, S., Morris, D., Radha, H.: Fast-clocs: Fast camera-lidar object candidates fusion for 3d object detection. In: WACV. pp. 187--196 (2022)

\bibitem{lss}
Philion, J., Fidler, S.: Lift, splat, shoot: Encoding images from arbitrary camera rigs by implicitly unprojecting to 3d. In: Computer Vision--ECCV 2020: 16th European Conference, Glasgow, UK, August 23--28, 2020, Proceedings, Part XIV 16. pp. 194--210. Springer (2020)

\bibitem{dreamfusion}
Poole, B., Jain, A., Barron, J.T., Mildenhall, B.: Dreamfusion: Text-to-3d using 2d diffusion. arXiv preprint arXiv:2209.14988  (2022)

\bibitem{f-pointnet}
Qi, C.R., Liu, W., Wu, C., Su, H., Guibas, L.J.: Frustum pointnets for 3d object detection from rgb-d data. In: CVPR. pp. 918--927 (2018)

\bibitem{pointnet}
Qi, C.R., Su, H., Mo, K., Guibas, L.J.: Pointnet: Deep learning on point sets for 3d classification and segmentation. In: CVPR. pp. 652--660 (2017)

\bibitem{complexer-yolo}
Simon, M., Amende, K., Kraus, A., Honer, J., Samann, T., Kaulbersch, H., Milz, S., Michael~Gross, H.: Complexer-yolo: Real-time 3d object detection and tracking on semantic point clouds. In: CVPR Workshops. pp.~0--0 (2019)

\bibitem{onecycle}
Smith, L.N., Topin, N.: Super-convergence: Very fast training of neural networks using large learning rates. In: Artificial intelligence and machine learning for multi-domain operations applications. vol. 11006, pp. 369--386. SPIE (2019)

\bibitem{ddim}
Song, J., Meng, C., Ermon, S.: Denoising diffusion implicit models. arXiv preprint arXiv:2010.02502  (2020)

\bibitem{rsn}
Sun, P., Wang, W., Chai, Y., Elsayed, G., Bewley, A., Zhang, X., Sminchisescu, C., Anguelov, D.: Rsn: Range sparse net for efficient, accurate lidar 3d object detection. In: CVPR. pp. 5725--5734 (2021)

\bibitem{efficientdet}
Tan, M., Pang, R., Le, Q.V.: Efficientdet: Scalable and efficient object detection. In: CVPR. pp. 10781--10790 (2020)

\bibitem{pointpainting}
Vora, S., Lang, A.H., Helou, B., Beijbom, O.: Pointpainting: Sequential fusion for 3d object detection. In: CVPR. pp. 4604--4612 (2020)

\bibitem{pointaugmenting}
Wang, C., Ma, C., Zhu, M., Yang, X.: Pointaugmenting: Cross-modal augmentation for 3d object detection. In: CVPR. pp. 11794--11803 (2021)

\bibitem{wang2022probabilistic}
Wang, T., Xinge, Z., Pang, J., Lin, D.: Probabilistic and geometric depth: Detecting objects in perspective. In: Conference on Robot Learning. pp. 1475--1485. PMLR (2022)

\bibitem{fcos3d}
Wang, T., Zhu, X., Pang, J., Lin, D.: Fcos3d: Fully convolutional one-stage monocular 3d object detection. In: ICCV. pp. 913--922 (2021)

\bibitem{detr3d}
Wang, Y., Guizilini, V.C., Zhang, T., Wang, Y., Zhao, H., Solomon, J.: Detr3d: 3d object detection from multi-view images via 3d-to-2d queries. In: Conference on Robot Learning. pp. 180--191. PMLR (2022)

\bibitem{m2bev}
Xie, E., Yu, Z., Zhou, D., Philion, J., Anandkumar, A., Fidler, S., Luo, P., Alvarez, J.M.: M$^2$bev: Multi-camera joint 3d detection and segmentation with unified birds-eye view representation. arXiv preprint arXiv:2204.05088  (2022)

\bibitem{fusionpainting}
Xu, S., Zhou, D., Fang, J., Yin, J., Bin, Z., Zhang, L.: Fusionpainting: Multimodal fusion with adaptive attention for 3d object detection. In: ITSC. pp. 3047--3054. IEEE (2021)

\bibitem{cmt}
Yan, J., Liu, Y., Sun, J., Jia, F., Li, S., Wang, T., Zhang, X.: Cross modal transformer: Towards fast and robust 3d object detection. In: ICCV. pp. 18268--18278 (2023)

\bibitem{second}
Yan, Y., Mao, Y., Li, B.: Second: Sparsely embedded convolutional detection. Sensors  \textbf{18}(10), ~3337 (2018)

\bibitem{deepinteraction}
Yang, Z., Chen, J., Miao, Z., Li, W., Zhu, X., Zhang, L.: Deepinteraction: 3d object detection via modality interaction. NeurIPS  \textbf{35},  1992--2005 (2022)

\bibitem{centerpoint}
Yin, T., Zhou, X., Krahenbuhl, P.: Center-based 3d object detection and tracking. In: CVPR. pp. 11784--11793 (2021)

\bibitem{mvp}
Yin, T., Zhou, X., Kr{\"a}henb{\"u}hl, P.: Multimodal virtual point 3d detection. NeurIPS  \textbf{34},  16494--16507 (2021)

\bibitem{deis}
Zhang, Q., Chen, Y.: Fast sampling of diffusion models with exponential integrator. arXiv preprint arXiv:2204.13902  (2022)

\bibitem{cvt}
Zhou, B., Kr{\"a}henb{\"u}hl, P.: Cross-view transformers for real-time map-view semantic segmentation. In: CVPR. pp. 13760--13769 (2022)

\bibitem{diff3det}
Zhou, X., Hou, J., Yao, T., Liang, D., Liu, Z., Zou, Z., Ye, X., Cheng, J., Bai, X.: Diffusion-based 3d object detection with random boxes. arXiv preprint arXiv:2309.02049  (2023)

\bibitem{voxelnet}
Zhou, Y., Tuzel, O.: Voxelnet: End-to-end learning for point cloud based 3d object detection. In: CVPR. pp. 4490--4499 (2018)

\bibitem{diffbev}
Zou, J., Zhu, Z., Ye, Y., Wang, X.: Diffbev: Conditional diffusion model for bird's eye view perception. arXiv preprint arXiv:2303.08333  (2023)

\end{thebibliography}

\clearpage
\setcounter{page}{1}
\maketitlesupplementary
\vspace{-2em}
\section{Additional implementation details}
\subsection{Formulation of Diffusion Models}
\label{appendix:diffusion_exposition}

Here, we review the formulation of diffusion models used in DiFFUSER, which based on the frameworks established in~\cite{ddpm, ddim, cfg}. Beginning with an original latent features $\bm{x}^F_0 \sim q(\bm{x}^F_0)$, we apply a sequential noising mechanism $q$, which incrementally infuses Gaussian noise into latent features $\bm{x}^F_1$, $\bm{x}^F_2$, ..., $\bm{x}^F_T$ at each time $t$. Specifically, the noise infusion at each stage is determined by a variance schedule $\beta_t \in (0, 1)$:
\begin{align}
    q(\bm{x}^F_{1:T}|\bm{x}^F_0) &= \prod^T_{t=1}q(\bm{x}^F_t | \bm{x}^F_{t-1}) \\
    q(\bm{x}^F_t | \bm{x}^F_{t-1}) &= \mathcal{N}(\bm{x}^F_t; \sqrt{1-\beta_t}\bm{x}^F_{t-1}, \beta_t \bm{I})
\end{align}

Following DDPM~\cite{ddpm}, instead of iteratively apply $q$, it is feasible to sample $\bm{x}^F_t$ at any chosen $t$ directly:

\begin{align}
    q(\bm{x}^F_t|\bm{x}^F_0) &= \mathcal{N}(\bm{x}^F_t; \sqrt{\bar{\alpha}_t}\bm{x}^F_0, (1 - \bar{\alpha}_t)\bm{I}) \\
    &= \sqrt{\bar{\alpha}_t} \bm{x}^F_0  + \epsilon \sqrt{1 - \bar{\alpha}_t}, \epsilon \sim \mathcal{N}(0, \bm{I})\label{eq:direct_sampling}
\end{align}
where $\bar{\alpha}_t = \prod_{s=0}^{t} \alpha_s$ and $\alpha_t = 1 - \beta_t$. In this context, $\bar{\alpha}_t$ is employed for scheduling the noise. According to Bayes' theorem, the posterior $q(\bm{x}^F_{t-1}|\bm{x}^F_t, \bm{x}^F_0)$ is also Gaussian:

\begin{align}
     q(\bm{x}^F_{t-1}|\bm{x}^F_t, \bm{x}^F_0) &= \mathcal{N}(\bm{x}^F_{t-1}; \mu_t'(\bm{x}^F_t, \bm{x}^F_0), \beta_t' \mathbf{I})\label{eq:reverse_posterior}
\end{align}

where the mean and variance, given by
\begin{align}
    \mu_t'(\bm{x}^F_t, \bm{x}^F_0) &=
    \frac{\sqrt{\bar{\alpha}_{t-1}}\beta_t}{1-\bar{\alpha}_t}\bm{x}^F_0 + \frac{\sqrt{\alpha_t}(1-\bar{\alpha}_{t-1})}{1-\bar{\alpha}_t} \bm{x}^F_t \label{eq:mean_tilde}
\end{align}
and
\begin{align}
    \beta_t' &= \frac{1-\bar{\alpha}_{t-1}}{1-\bar{\alpha}_t} \beta_t \label{eq:variance_tilde}
\end{align}
respectively, characterize this Gaussian distribution.

To generate a sample from $q(\bm{x}^F_0)$, we may sample from $q(\bm{x}^F_T)$ and iteratively apply the reverse transitions $q(\bm{x}^F_{t-1} | \bm{x}^F_t)$ back to $\bm{x}^F_0$. Given a sufficiently extensive $T$ and a suitably chosen $\beta_t$ series ($\beta_t \rightarrow 0$), the distribution $q(\bm{x}^F_T)$ converges closely to an isotropic Gaussian. Specifically, our conditional diffusion model is parametrized through $p_{\theta}(x^F_{t-1}|x^F_t, t,\tilde{x}^F_0)$. Because $p_{\theta}(x^F_{t-1}|x^F_t, t,\tilde{x}^F_0)$ is not be easily estimated but could be approximated using a neural network (~\ie DifFUSER), where the aim is to predict $\mu_\theta$ and $\Sigma_\theta$:
\begin{align}
p_{\theta}(\bm{x}^F_{t-1}|x^F_t, t,\tilde{x}^F_0) &\coloneqq \mathcal{N}(\bm{x}^F_{t-1};\mu_{\theta}(x^F_t, t,\tilde{x}^F_0), \Sigma_{\theta}(x^F_t, t,\tilde{x}^F_0)) \label{eq:ptheta}
\end{align}

Here, a neural network can also be optimized to predict either $\epsilon$ or $\bm{x}^F_0$ \cite{ddim, cfg}. We opt for predicting $\bm{x}^F_0$ in this paper.
\clearpage

\subsection{Pseudo-code of DifFUSER training and sampling}
\begin{table}[ht]
\vspace{-4em}
\centering
\begin{algorithm}[H]
\small
\caption{DifFUSER Training}
\begin{algorithmic}
\ttfamily
\small
\Function{train\_loss}{images, pointcloud, gt\_boxes, gt\_map}
    \State \textcolor{OliveGreen}{"""}
    \State \textcolor{OliveGreen}{Input: }
    \State \hspace{1em} \textcolor{OliveGreen}{images: [B, V, H, W, 3] \# V: number of multi-view images}
    \State \hspace{1em} \textcolor{OliveGreen}{pointcloud: [B, N, 3] \# N: number of points in pointcloud}
    \State \hspace{1em} \textcolor{OliveGreen}{gt\_boxes: [B, *, D] \# D: number of box dimensions}
    \State \hspace{1em} \textcolor{OliveGreen}{gt\_map: [B, C, 200, 200] \# C: number of segmentation classes}
    \State \textcolor{OliveGreen}{Output: total loss which is a measure of the prediction quality}
    \State \textcolor{OliveGreen}{"""}
    \State \textcolor{OliveGreen}{\# Encode the image and LiDAR pointcloud features}
    \State cam\_feats $\gets$ \text{image\_encoder}(images)
    \State lidar\_feats $\gets$ \text{lidar\_encoder}(pointcloud)
    \State 

    \State \textcolor{OliveGreen}{\# Concatenate extracted features forming initial latent}
    \State \textcolor{OliveGreen}{\# representation and set as diffusion target}
    \State $x^F_0 \gets$ \text{concatenate}(cam\_feats, lidar\_feats)
    \State diffusion\_target $\gets x^F_0$
    \State 
    
    \State \textcolor{OliveGreen}{\# Perturb the initial representation using PDST}
    \State $\tilde{x}^F_0 \gets \text{PDST}(x^F_0)$
    
    \State \textcolor{OliveGreen}{\# Randomly select a time step in the diffusion process}
    \State $t \gets \text{randint}(0, T)$
    
    \State noise $\gets \text{normal}(\text{mean=0, std=1})$ \textcolor{OliveGreen}{\# Generate noise}
    
    \State \textcolor{OliveGreen}{\# Apply the noise to the latent representation at time step t}
    \State $x^F_t$ $\gets$ sqrt(alpha\_cumprod(t) * $x^F_0$ + sqrt(1-alpha\_cumprod(t)) * noise
    
    \State \textcolor{OliveGreen}{\# Predict the denoised features at time t using the DifFUSER model}
    \State $\hat{x}^F_0$ $\gets$ DifFUSER$(\tilde{x}^F_0, t)$
    \State \textcolor{OliveGreen}{\# Pass the enhanced features through the task-specific head}
    \State pred $\gets$ task\_head($\hat{x}^F_0$)

    \State \textcolor{OliveGreen}{\# Compute the total loss}
    \State loss $\gets$ diffusion\_loss(diffusion\_target, $\hat{x}^F_0$)
    \Statex \hspace{1.6cm} $+$ task\_loss(pred, gt\_boxes, gt\_map)
    
    \State \Return loss
\EndFunction
\end{algorithmic}
\end{algorithm}
\end{table}

\clearpage
\begin{table}[ht]
\centering
\begin{algorithm}[H]
\small
\caption{DifFUSER Sampling}
\begin{algorithmic}
\ttfamily
\small
\Function{train\_loss}{images, pointcloud, steps, T}
    \State \textcolor{OliveGreen}{"""}
    \State \textcolor{OliveGreen}{Input: }
    \State \hspace{1em} \textcolor{OliveGreen}{images: [B, V, H, W, 3] \# V: number of multi-view images}
    \State \hspace{1em} \textcolor{OliveGreen}{pointcloud: [B, N, 3] \# N: number of points in pointcloud}
    \State \hspace{1em} \textcolor{OliveGreen}{steps: number of sampling steps}
    \State \hspace{1em} \textcolor{OliveGreen}{T: number of time steps}
    \State \textcolor{OliveGreen}{Output: task-specific predictions}
    \State \textcolor{OliveGreen}{"""}
    \State \textcolor{OliveGreen}{\# Encode the image and LiDAR pointcloud features}
    \State cam\_feats $\gets$ \text{image\_encoder}(images)
    \State lidar\_feats $\gets$ \text{lidar\_encoder}(pointcloud)
    \State 

    \State \textcolor{OliveGreen}{\# Concatenate extracted features forming initial latent}
    \State \textcolor{OliveGreen}{\# representation and use as diffusion condition}
    \State $x^F_0 \gets$ \text{concatenate}(cam\_feats, lidar\_feats)
    \State 
    
    \State \textcolor{OliveGreen}{\# Sampling step sizes [(T-1, T-2), (T-2, T-3), ..., (1, 0)]}
    \State times $\gets$ reverse(linspace(-1, T - 1, steps + 1))
    \State time\_pairs $\gets$ list(zip(times[:-1], times[1:]))
    
    \State
    \State $x^F_t$ $\gets \text{normal}(\text{mean=0, std=1})$ \textcolor{OliveGreen}{\# Sample $x^F_t$ from noise}
    
    \State for (t\_now, t\_next) in time\_pairs:
    \State \hspace{1em} $\hat{x}^F_0 \gets $ DifFUSER($x^F_0, x^F_t,$ t\_now)
    \State \textcolor{OliveGreen}{\# estimate $x^F_t$ at t\_next, sampler can be \cite{ddim}, \cite{deis}, or \cite{dpmsolver++}}
    \State \hspace{1em} $x^F_t \gets $ sampler\_step($\hat{x}^F_0$, t\_now, t\_next) 

    \State \textcolor{OliveGreen}{\# Pass the predicted enhanced features through the task-specific head}
    \State pred $\gets$ task\_head($\hat{x}^F_0$)
    
    \State \Return pred
\EndFunction
\end{algorithmic}
\end{algorithm}
\end{table}
\clearpage



\section{Additional Qualitative Results}
\includegraphics[width=\textwidth]{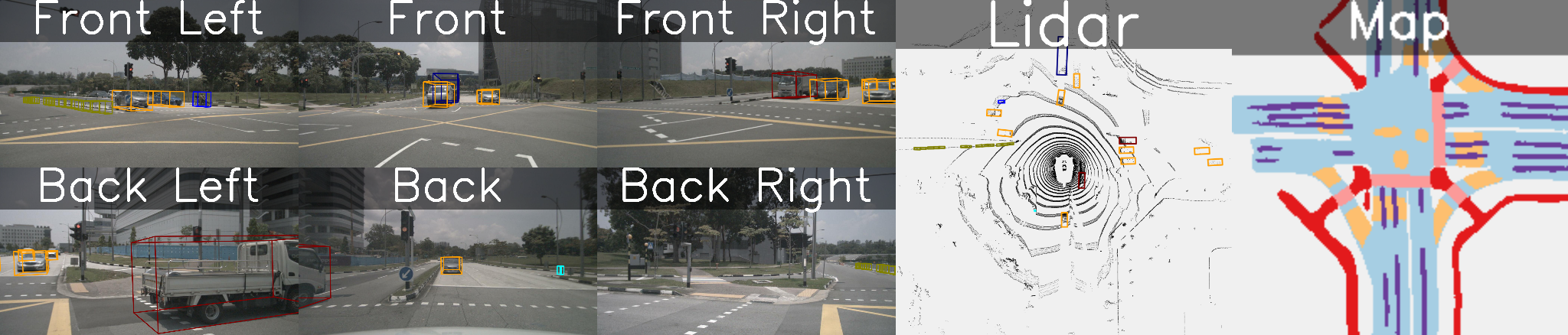} 
\includegraphics[width=\textwidth]{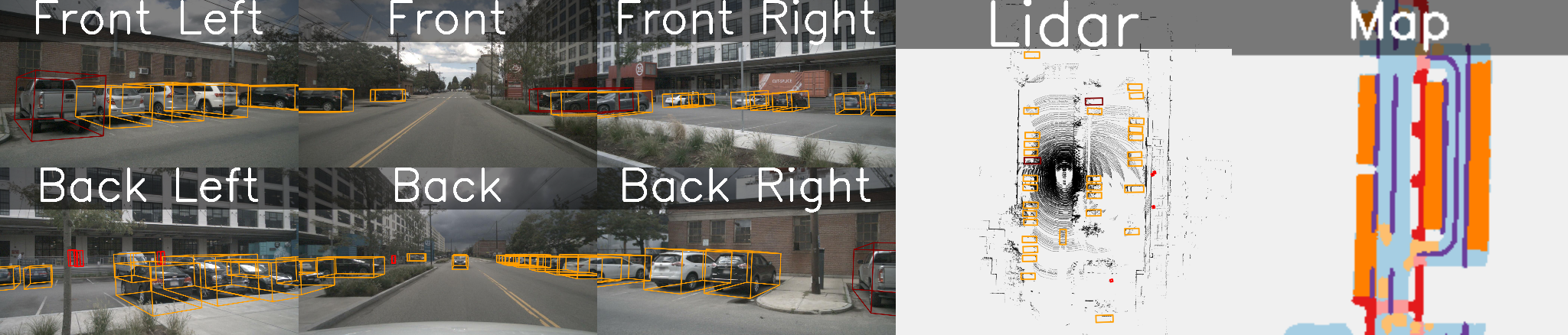} 
\includegraphics[width=\textwidth]{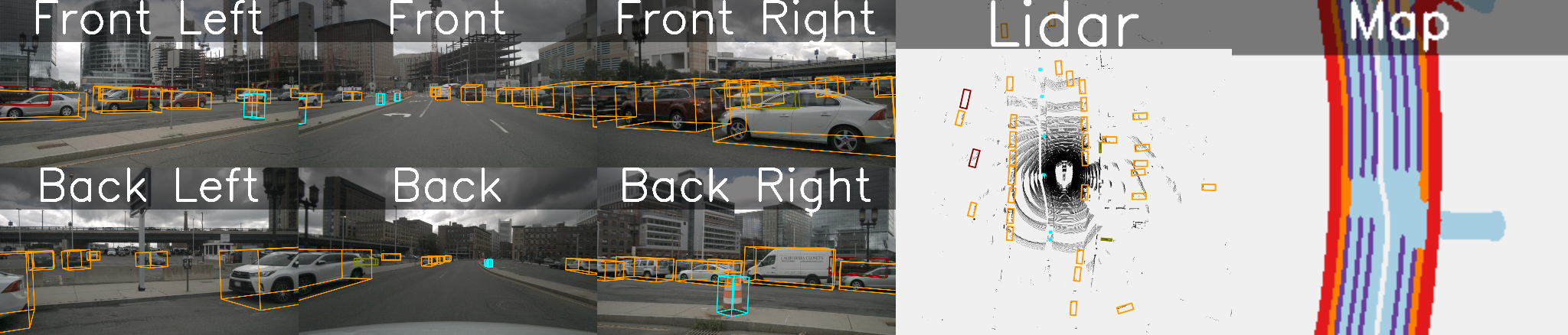} 
\includegraphics[width=\textwidth]{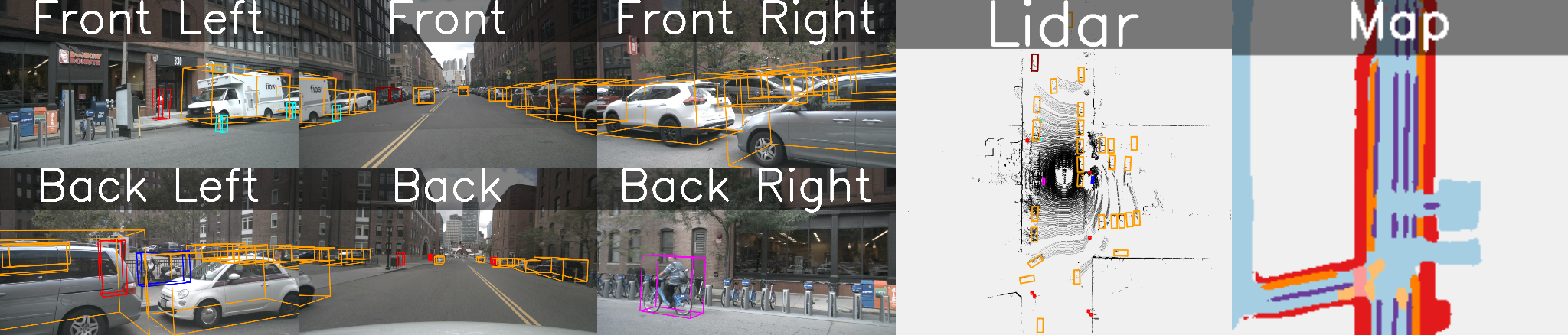} 
\includegraphics[width=\textwidth]{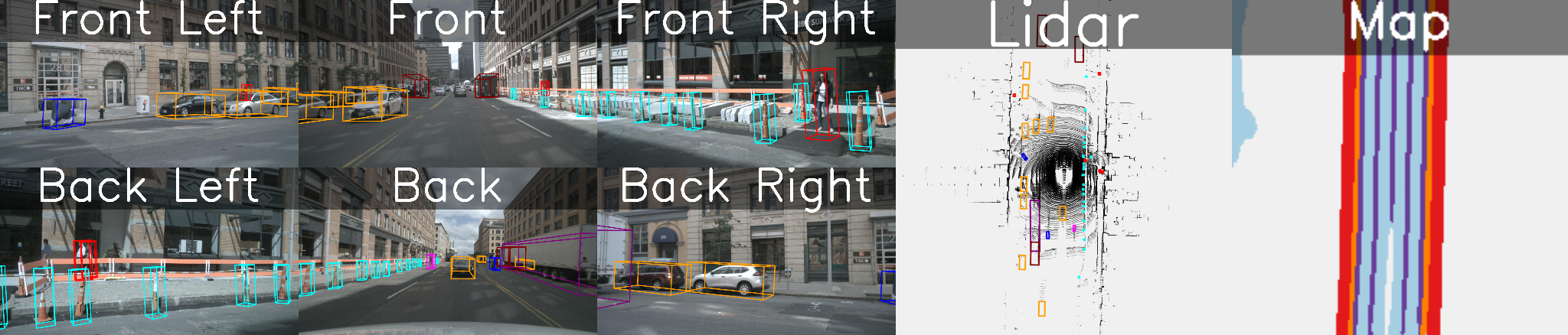}

\clearpage
\begin{figure*}[h]
    \centering
   \includegraphics[width=\linewidth]{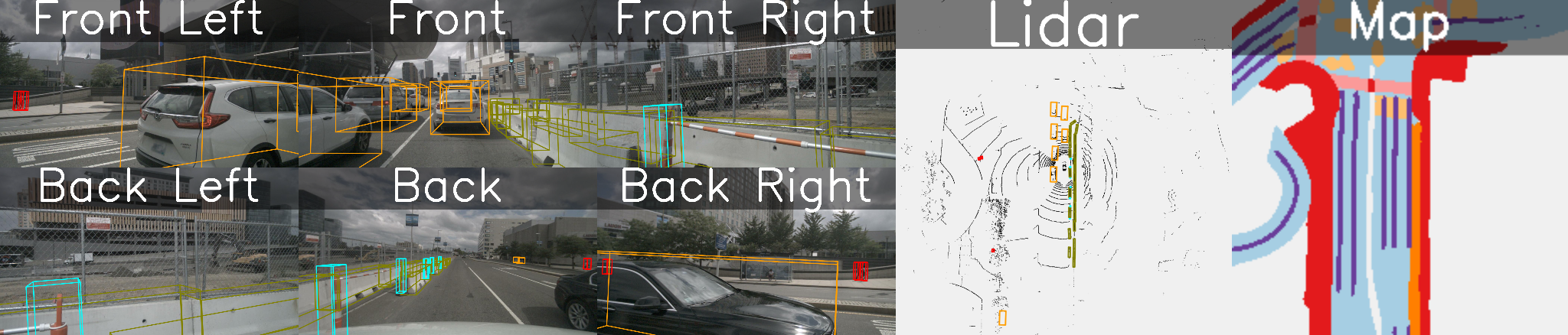}
   \includegraphics[width=\linewidth]{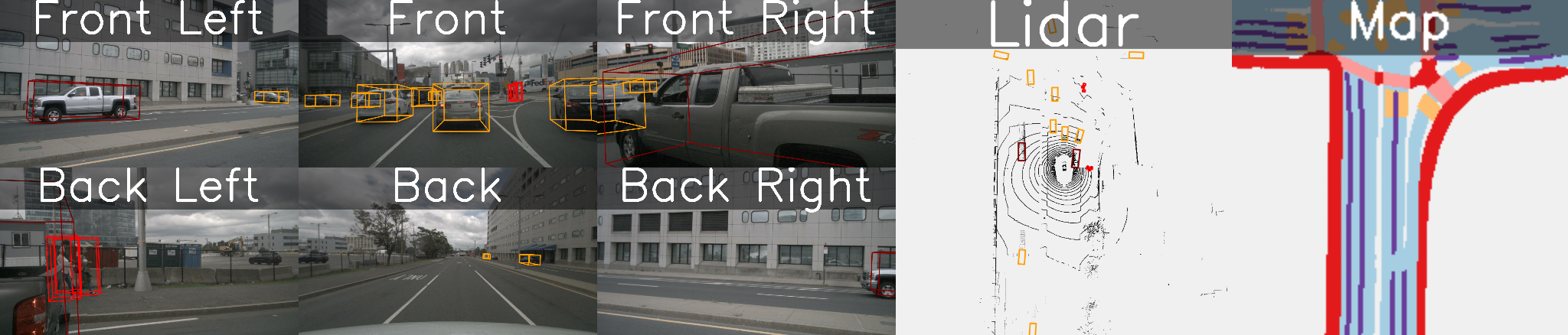}
   \includegraphics[width=\linewidth]{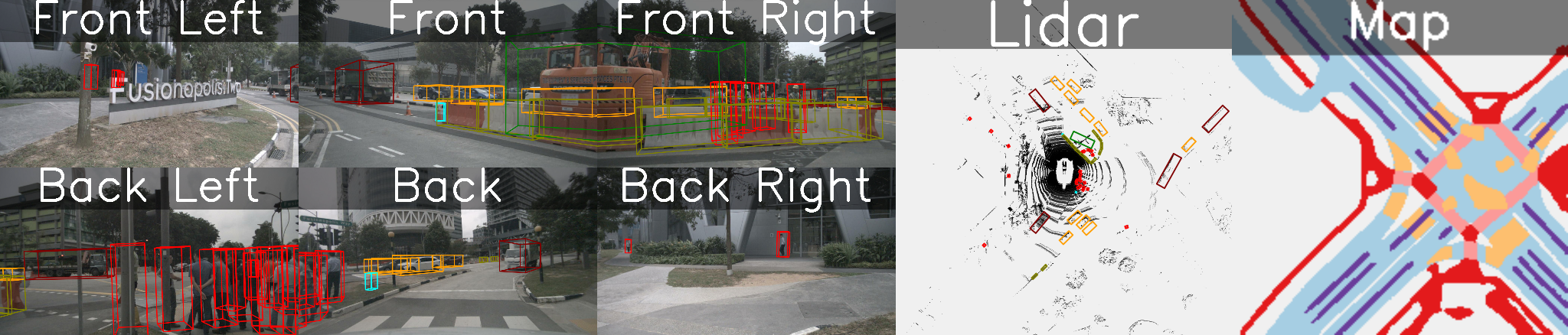}
   \includegraphics[width=\linewidth]{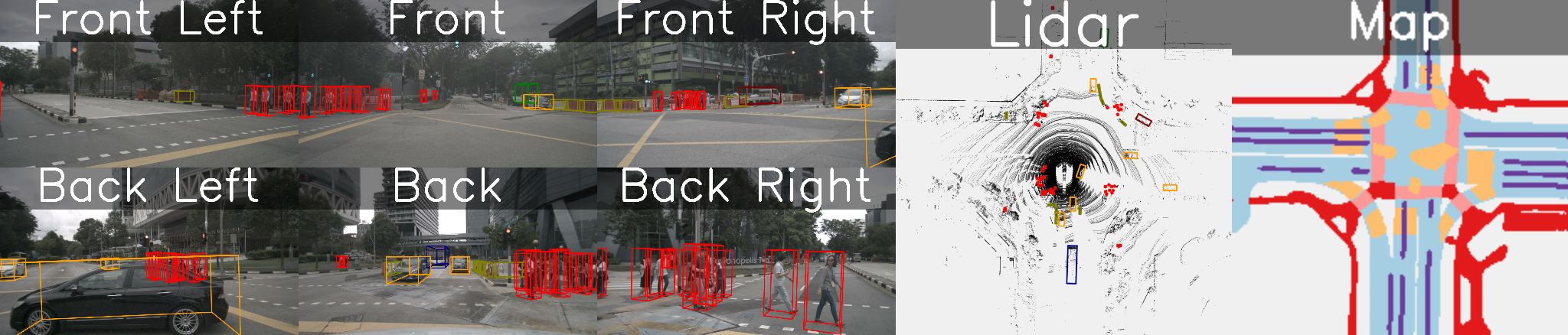}
   \includegraphics[width=\linewidth]{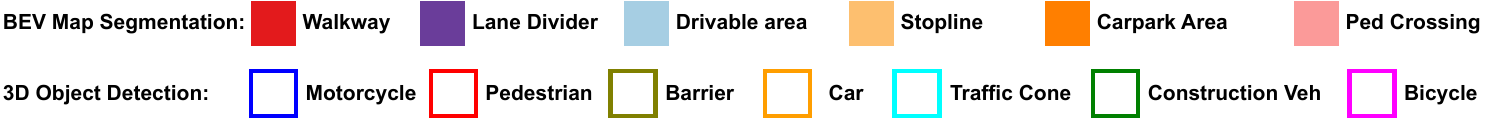}
    \caption{We provide additional qualitative results for challenging scenarios from the NuScenes \textit{val} set. On the left, we show the projections of predicted 3D bounding boxes on the camera images. On the right, we show the 3D prediction (in BEV) and BEV map segmentation results. Best viewed in color and zoomed in.}
\end{figure*}

\end{document}